\documentclass[11pt]{article}
\usepackage[preprint]{acl}
\usepackage{times}
\usepackage{latexsym}
\usepackage[T1]{fontenc}
\usepackage[utf8]{inputenc}
\usepackage{microtype}
\IfFileExists{inconsolata.sty}{\usepackage{inconsolata}}{}
\usepackage{booktabs}
\usepackage{graphicx}
\usepackage{xcolor}
\usepackage{amsmath,amssymb}
\usepackage{multirow}
\usepackage{listings}
\usepackage{comment}
\usepackage{array}
\usepackage{fvextra}
\usepackage[normalem]{ulem}
\usepackage{xcolor}
\usepackage{url}

\definecolor{new}{HTML}{C45A00}

\newcommand{\resultfigure}[1]{\IfFileExists{#1}{\includegraphics[width=\columnwidth]{#1}}{\fbox{\parbox{0.88\columnwidth}{\centering\small Figure asset not supplied.\par\texttt{\detokenize{#1}}}}}}
\title{Beyond Hard Labels: LLM Probabilities Improve Human-Assisted Regression}
\author{
  Xinyue Wang \and Qunhua Li \\
  Department of Statistics \\
  The Pennsylvania State University \\
  University Park, PA 16802, USA \\
  \texttt{\{xpw5228, qunhua.li\}@psu.edu}
}
\begin{document}
\maketitle

\begin{abstract}
Large language models (LLMs) are increasingly used for text
annotation, but their outputs are often reduced to a single
categorical label before downstream analysis. This hardening step
discards the model's relative support across categories, which may
contain useful information when category boundaries are ambiguous.
We study whether retaining the full probability vector improves
downstream regression when LLM annotations are combined with a
limited amount of human annotation. Using matched soft and hard
representations from three LLMs across three social-science text
datasets, we compare direct regression and human-assisted
correction. Soft inputs are not uniformly better when used
directly as outcomes, but after human correction, using the
probability vector consistently improves regression efficiency
across models and datasets. Analyses of repeated human judgments
and controlled simulations further show that these gains arise
when probability variation beyond the hard category carries useful
information about the human-coded outcome. Our results identify
output representation as an important design choice in
LLM-assisted measurement and support preserving probability
outputs for downstream statistical analysis. Code and reproducibility materials are available at \url{https://anonymous.4open.science/r/soft-label-LLM-D201/}. 
\end{abstract}

\section{Introduction}
\label{sec:introduction}

Large language models (LLMs) are increasingly used to annotate
text at scales that would be costly to achieve with human coding
alone \cite{gilardi2023annotation}. For example, researchers may classify news sentences as factual versus opinionated, online
comments as hate speech versus non-hate speech, or event reports
by protest type.  In many social-science
applications, however, classification is not the endpoint. The
resulting annotations become measurements used in downstream
statistical analyses, for example to estimate how political
events, media content, or harmful speech vary across sources,
platforms, or time. In these settings, classification accuracy
alone is not sufficient: even accurate automated labels can yield
biased regression estimates when annotation errors are related to
the covariates of interest
\cite{grimmer2013text,egami2023dsl}.

Categorical annotation itself is often uncertain. Texts near the
boundary between categories may plausibly support more than one
interpretation, and human annotators frequently disagree on such
cases \cite{plank2022problem,uma2021disagreement}. A hard label
reduces this uncertainty to a single choice: it records which
category is selected, but not how strongly that category is
preferred over the alternatives. A probability vector preserves
this additional structure by expressing relative support across
categories. Elicited LLM probabilities need not be perfectly
calibrated to human judgments to be useful; variation across
categories may still carry information about the human-coded
outcome that is lost when only the most likely category is
retained. Prior work on soft labels and annotation distributions
has shown that richer representations can preserve useful
information beyond a single categorical decision
\cite{fornaciari-etal-2021-beyond,wu-etal-2023-dont,
baan-etal-2024-interpreting}.

Human-assisted inference provides a way to use automated
annotations without treating them as error-free measurements.
Prediction-powered inference (PPI) and design-based supervised
learning (DSL), for example, combine model-generated measurements
for a large corpus with human labels for a smaller sample
\cite{angelopoulos2023ppi,egami2023dsl}. These approaches show
how imperfect predictions can support downstream inference, but
largely take the representation of those predictions as given.
Yet representation is itself a measurement-design choice. If a
probability vector contains information relevant to the
human-coded outcome, reducing it to a hard label discards that
information before correction or downstream analysis begins.

We study this question by comparing the full category-probability
output from an LLM with the hard label derived from the same model
response. This paired comparison holds the underlying prediction
fixed and changes only how much of the model output is retained.
We ask whether the information discarded by hardening is related
to human judgments, whether retaining it improves downstream
regression when limited human annotation is used for correction,
and whether those gains translate into lower human-annotation
requirements. We examine these questions using repeated human
judgments, direct and learned human-assisted correction,
annotation-budget analyses, and controlled simulations.

Our results show that retaining the probability vector preserves
information that is relevant to human judgments and can improve
the efficiency of human-assisted regression. More broadly, the
findings highlight output representation as an important part of
LLM-assisted measurement: downstream performance depends not only
on which category a model selects, but also on what information
from the model output is retained for subsequent analysis.


\section{Related Work}
\label{sec:related-work}

\subsection{Human Label Variation and Soft Labels}

Human annotators often disagree, and a growing NLP literature
argues that collapsing repeated judgments to a single majority
label can discard meaningful structure
\cite{uma2021disagreement,plank2022problem,leonardelli2023lewidi}.
Prior work has therefore represented annotation variation through
label distributions, annotator confidence, and secondary labels,
and has used these richer signals as soft supervision for model
training \cite{fornaciari-etal-2021-beyond,wu-etal-2023-dont}.
Related work has also examined how predictive probability
distributions relate to human label variation
\cite{baan-etal-2024-interpreting}. More recently, studies of
LLM-based annotation have asked whether distributions induced by
model annotations reproduce patterns of human disagreement,
finding partial alignment as well as model-specific differences
\cite{kulmizev-etal-2026-label}.

Together, this literature shows that information beyond a single
categorical label can be substantively and statistically
meaningful. We build on this perspective by studying the
downstream inferential value of probability information produced
by an LLM annotator.

\subsection{LLMs as Text Annotators and Confidence Elicitation}

LLMs are increasingly used as scalable annotators, including in
computational social science
\cite{gilardi2023annotation,ziems-etal-2024-large}. This work has
primarily evaluated categorical accuracy, agreement with human
annotators, and sensitivity to task instructions and coding
schemes. Recent studies also show that models can depart from
researcher-specified codebook definitions even when their final
predictions appear plausible
\cite{halterman2026codebook}.

A related literature studies confidence and probability
elicitation from language models. Verbalized or categorical
probabilities can contain useful information about prediction
quality, although their calibration varies across models, tasks,
prompts, and elicitation procedures
\cite{tian2023calibration,xiong2024uncertainty}. These studies
highlight that LLM outputs can carry information beyond the
selected class itself.

Our focus is downstream: when an LLM is used as an annotator,
does retaining the full category-probability output rather than
only its most likely category improve subsequent statistical
analysis?

\subsection{Inference with Predicted Measurements}

Machine-generated annotations can introduce systematic error when
they are subsequently used as variables in statistical analysis.
Post-prediction inference has shown that treating predicted
outcomes as observed can bias regression coefficients and
uncertainty estimates \cite{wang2020postprediction}. Related work
in automated content analysis similarly demonstrates how
classification error can propagate into downstream regression
estimates \cite{teblunthuis2024misclassification}.

Recent methods address this problem by combining large-scale
predictions with a smaller set of human-reference observations.
Prediction-powered inference (PPI) uses labeled data to correct
inference based on model predictions
\cite{angelopoulos2023ppi}, with later work improving statistical
efficiency \cite{angelopoulos2026ppi++}. Design-based supervised
learning (DSL) develops related correction procedures for
probability-sampled labels and automated text measurements used in
downstream regression \cite{egami2023dsl}. These methods establish
that imperfect automated measurements can still support reliable
and efficient inference when anchored by human labels.

We extend this line of work by showing that the representation
retained from an LLM annotation---the full probability vector
versus its argmax---can itself materially affect the efficiency
of human-assisted downstream inference.

\section{Methods}
\label{sec:method}

\subsection{Setup and Target}
\label{sec:representations}
\label{sec:target}

Consider a corpus of $N$ texts and a categorical annotation task with $K\geq2$ mutually exclusive categories. For text $i$, let $C_i\in\{1,\ldots,K\}$ denote the human-reference category and
\[
Y_{ik}=\mathbb{I}(C_i=k)
\]
its indicator for category $k$. The construction of the human-reference category for each dataset is specified in Section~\ref{sec:datasets}.

For the same categories, an LLM returns a probability vector
\[
Q_i=(Q_{i1},\ldots,Q_{iK})^\top,\quad
Q_{ik}\geq0, \sum_{k=1}^K Q_{ik}=1.
\]
We call $Q_i$ the \emph{soft} representation. Let
\[
\widehat C_i=\arg\max_k Q_{ik}
\]
denote the LLM's hard predicted category, and let
\[
H_i=\operatorname{onehot}(\widehat C_i)
\]
denote its matched \emph{hard} one-hot representation. Because both $\widehat C_i$ and $H_i$ are constructed deterministically from $Q_i$, each soft--hard comparison holds the underlying LLM output fixed and differs only in whether information beyond the argmax is retained. We treat $Q_i$ as a model-generated representation and do not assume that it is calibrated to human judgments.

Let $x_i\in\mathbb{R}^p$ contain downstream covariates, including an intercept, and let $X$ denote the corresponding $N\times p$ design matrix. For category $k$, the target is the regression coefficient
\begin{equation}
\beta_k=(X^\top X)^{-1}X^\top Y_{\cdot k}.
\label{eq:regression-target}
\end{equation}
Thus, the LLM annotation is an intermediate measurement: the estimand is defined by the human-reference outcome and its relationship with the downstream covariates.

\subsection{Human-Assisted Regression}
\label{sec:human-assisted}

We build on prediction-plus-residual estimators used in
prediction-powered inference and design-based supervised learning
\cite{angelopoulos2023ppi,egami2023dsl}. Suppose human outcomes
are observed for a probability sample of texts. Let $S_i$ indicate
whether text $i$ is sampled for human annotation and let
$\pi_i>0$ denote its inclusion probability. For any auxiliary
prediction $g_{ik}$ of $Y_{ik}$, define the pseudo-outcome
\begin{equation}
\widetilde{Y}_{ik}(g)
=
g_{ik}
+
\frac{S_i}{\pi_i}
\left(Y_{ik}-g_{ik}\right).
\label{eq:pseudo-outcome}
\end{equation}
We then estimate
\begin{equation}
\widehat\beta_k(g)
=
(X^\top X)^{-1}X^\top
\widetilde{Y}_{\cdot k}(g).
\label{eq:corrected-regression}
\end{equation}

The first term in Equation~\eqref{eq:pseudo-outcome} uses auxiliary predictions available throughout the corpus, while the second uses sampled human outcomes to correct their discrepancies from the human-reference target. This formulation separates the outcome being estimated, which remains human defined, from the auxiliary representation used to improve estimation. Sampling-design and implementation details are given in Appendix~\ref{sec:correction-implementation}, with variance estimation in Appendix~\ref{sec:variance-details}.

\subsection{Soft and Hard Estimation Strategies}
\label{sec:estimators}

We compare $Q_i$ and its matched hard reduction $H_i$ in three settings.

\paragraph{Uncorrected regression.}
Without human correction, the chosen LLM representation directly replaces the human outcome:
\[
\widehat\beta_{Q,k}=(X^\top X)^{-1}X^\top Q_{\cdot k},
\]
\[
\widehat\beta_{H,k}=(X^\top X)^{-1}X^\top H_{\cdot k}.
\]
This comparison asks whether the numerical probability vector is a better direct measurement than its hard reduction without sampled human labels.

\paragraph{Direct correction.}
Following the prediction-plus-residual idea in PPI \cite{angelopoulos2023ppi}, we use the LLM output itself as the auxiliary prediction in Equations~\eqref{eq:pseudo-outcome}--\eqref{eq:corrected-regression}:
\[
g_{ik}=Q_{ik}\quad\text{(soft)},\qquad
g_{ik}=H_{ik}\quad\text{(hard)}.
\]
We refer to these as PPI-style soft and hard correction. Neither fits an additional auxiliary model.

\paragraph{Learned correction.}
Following DSL \cite{egami2023dsl}, we instead learn the auxiliary prediction from sampled human labels. Soft+DSL uses a category-specific model $g_k^Q(x_i,Q_i)$, whereas DSL uses the matched model $g_k^H(x_i,H_i)$. The two procedures use the same human-label sample, downstream covariates, folds, and auxiliary-model family; the only difference is whether the model receives $Q_i$ or $H_i$. Our primary implementation uses linear auxiliary models with cross-fitting. Full implementation details are given in Appendices~\ref{sec:correction-implementation} and~\ref{sec:cross-fitting}.

\subsection{Representation and Statistical Efficiency}
\label{sec:efficiency}

Human labels define the target, while the auxiliary prediction
determines how much information can be borrowed from the rest of
the corpus. For a fixed $g_{ik}$,
\[
\mathbb{E}_S\!\left[
g_{ik}+\frac{S_i}{\pi_i}(Y_{ik}-g_{ik})
\right]
=Y_{ik},
\]
so the correction remains anchored to the human-reference outcome.
The quality of $g_{ik}$ mainly affects precision: a more informative
auxiliary prediction leaves smaller residuals $Y_{ik}-g_{ik}$ to
be recovered from the limited human-labeled sample. We use
\emph{statistical efficiency} to mean lower estimation error or
narrower intervals for a fixed human-label budget.

This makes the representation supplied to $g$ consequential.
The hard label retains only the most likely category, whereas the
probability vector also preserves variation in relative support
among categories. Because the hard representation is a
deterministic reduction of the probability vector, soft inputs can
improve efficiency only when this additional variation helps
predict the human-reference outcome. This applies both when the
LLM output is used directly as $g$ and when it enters a learned
auxiliary model. Conversely, if the probability vector contains
no information beyond the hard category, the two representations
should offer no systematic efficiency advantage. We examine both
possibilities empirically and with a controlled negative
simulation.

\section{Experimental Setup}
\label{sec:experiments}
\subsection{Datasets and Downstream Regression Specifications}
\label{sec:datasets}
We evaluate three social-science text datasets: BABE opinion annotations \cite{spinde2021babe}, Measuring Hate Speech (MHS) \cite{sachdeva2022mhs,kennedy2020mhs}, and Crowd Counting Consortium (CCC) event data \cite{halterman2026codebook,halterman2025replication,ccc2017data}. Table~\ref{tab:dataset-design} summarizes corpus sizes, downstream contrasts, and primary annotation budgets. BABE and MHS use fixed plurality human-reference categories; CCC uses its archived event category. Repeated judgments in BABE and MHS are used for descriptive vote-distribution analyses but do not replace the categorical regression outcome. Primary analyses use texts with valid probability outputs from all three models. Full dataset preparation and regression specifications appear in Appendix~\ref{sec:data-details}.

\begin{table*}[t]
\centering
\small
\setlength{\tabcolsep}{5pt}
\renewcommand{\arraystretch}{1.05}
\begin{tabular}{@{}lcll@{}}
\toprule
Dataset & Corpus & Outcome / downstream contrasts & Primary human budget \\
\midrule
BABE & 3,575 & Opinion; left/right vs.\ center (6 slopes) & 500 judgments (100 $\times$ 5) \\
MHS  & 2,974 & Hate speech; Twitter/YouTube vs.\ Reddit (6 slopes) & 300 comments \\
CCC  & 4,271 & Event type; year + region contrasts (24 slopes) & 450 labels \\
\bottomrule
\end{tabular}
\caption{Empirical datasets, downstream contrasts, and primary human-label budgets.}
\label{tab:dataset-design}
\end{table*}

\subsection{LLMs and Probability Elicitation}
\label{sec:probability-elicitation}
The primary analysis compares GPT-5.6 Luna \cite{openai2026gpt56}, Claude Sonnet~5 \cite{anthropic2026sonnet5}, and Qwen3-32B \cite{yang2025qwen3}. Each model receives the source text and the task's coding instructions in a zero-shot prompt requesting category probabilities with at most four decimal places and no explanation. Human labels, downstream covariate fields, labeled examples, and retrieved context are not appended. Additional reasoning or thinking is disabled for these runs. The source text itself may contain information related to the downstream covariates.

Each soft--hard pair is derived from the same validated probability output; no separate hard-label prompt is used. Output validation, complete prompts, model identifiers, and provider-specific settings are given in Appendices~\ref{sec:prompt-details} and~\ref{sec:full-prompts}.

\subsection{Human-Annotation Budgets and Evaluation Protocol}
\label{sec:label-sampling}
Budgets count human annotations made available to an estimator. Selecting a BABE sentence reveals all five judgments; selecting an MHS comment reveals all existing judgments; CCC uses one archived event label per selected event. 

Each primary setting uses 2,000 paired repetitions of human-label sampling. The observed corpus and LLM outputs remain fixed, and paired soft--hard methods use identical sampled labels and fold assignments. We also include a same-budget human-only reference that uses the sampled human labels and full-corpus covariates but no LLM output; its implementation is given in Appendix~\ref{sec:cross-fitting}. Complete sampling designs, strata, and tested budget grids are given in Appendix~\ref{sec:sampling-details}.

The primary endpoint is aggregate coefficient RMSE, with squared errors pooled equally across the designated slopes and repetitions before taking the square root. We also report relative soft--hard RMSE reduction and evaluate nominal 95\% pointwise confidence intervals by their mean coverage and width. RMSE and interval width are expressed in percentage points. Empirical intervals quantify uncertainty from revealing only a subset of the corpus's human labels. Metric definitions, coefficient-bias diagnostics, and empirical and simulation variance calculations are given in Appendix~\ref{sec:variance-details}. Section~\ref{sec:annotation-equivalent} compares annotation budgets needed to achieve matched coefficient RMSE.

\section{Results}
\label{sec:results}

We organize the results around three questions. We first ask whether the information discarded by hardening is related to human judgments, then whether retaining that information improves human-assisted regression, and finally how those gains translate into human-annotation requirements. We conclude with mechanism and robustness checks.

\subsection{Human-Relevant Information Beyond the Hard Label}
\label{sec:information-beyond-hard}

\begin{figure}[t]
    \centering
    \includegraphics[width=1\linewidth]{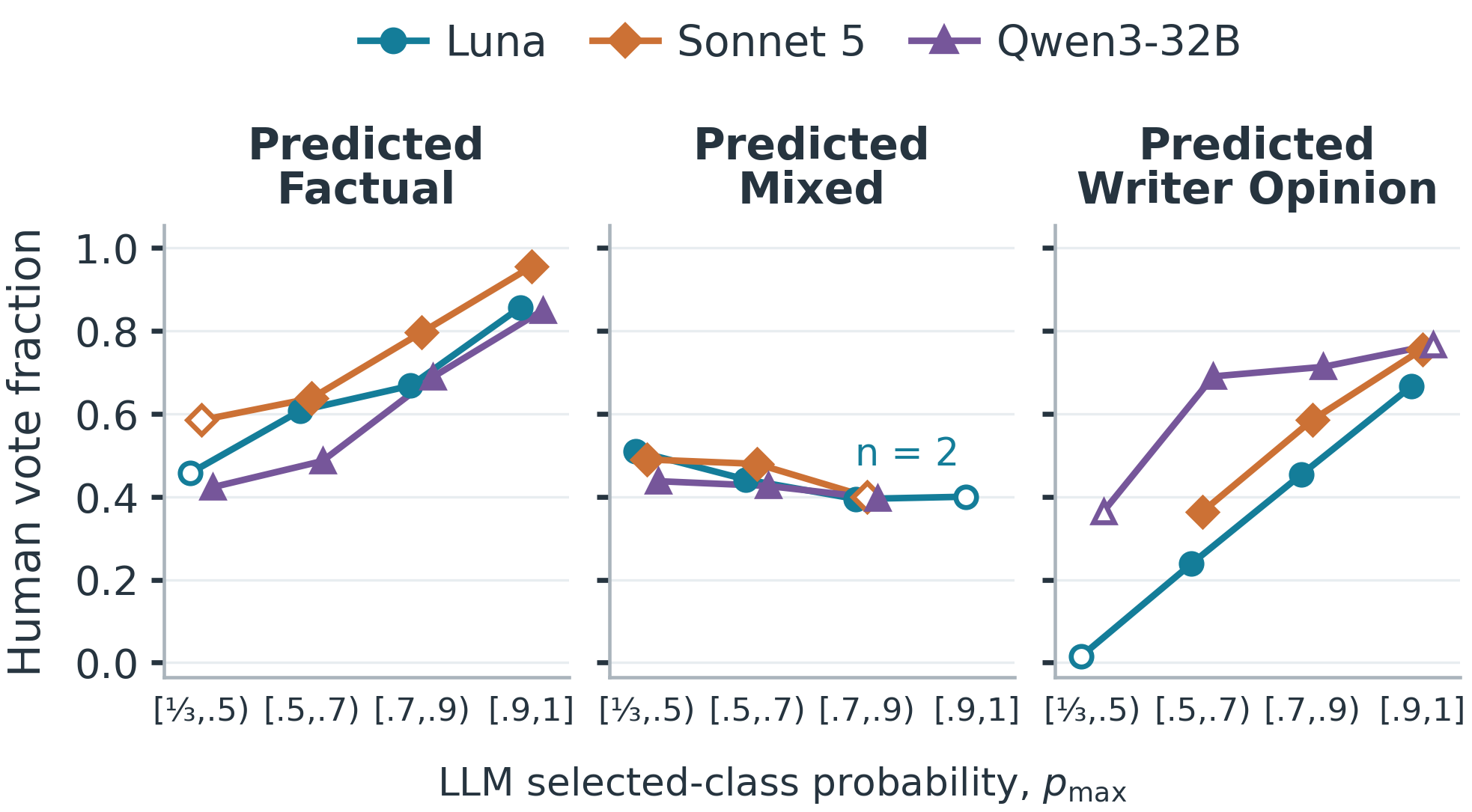}
    \caption{Human vote alignment within fixed LLM hard predictions in BABE. Within each predicted category, the x-axis bins $p_{\max}$ and the y-axis gives the mean fraction of five human annotators selecting that category. Open markers denote bins with fewer than 20 texts.}
    \label{fig:within-hard-babe}
\end{figure}

The central representation question is whether variation discarded by the argmax is related to human judgments. BABE provides a direct diagnostic because each text has five human annotations. Let $R_i$ denote the empirical human vote proportions for text $i$. For each LLM, define
$
A_i = R_{i,\widehat C_i},
$
the fraction of human annotators selecting the LLM's hard predicted category, and let $p_{\max,i}=Q_{i,\widehat C_i}=\max_k Q_{ik}$ denote the probability assigned to that category. We group texts by $p_{\max}$ and compute mean $A_i$ separately within each LLM-predicted category. 
Because the hard predicted category is fixed within each panel of Figure~\ref{fig:within-hard-babe}, an association between $p_{\max}$ and mean human agreement indicates that the soft probabilities contain human-relevant information beyond the hard label.

Figure~\ref{fig:within-hard-babe} shows a clear positive relationship for texts predicted as \textsc{Factual} and \textsc{Writer Opinion}: across all three LLMs, larger selected-class probabilities are generally associated with larger fractions of human votes for the same category. Although no comparable trend is evident for \textsc{Mixed}, these results show that selected-class probabilities can provide information about human judgments even among texts receiving the same hard prediction.

A complementary distribution-level comparison gives the same overall direction. We compare the soft vector $Q_i$ and its matched hard one-hot vector $H_i$ with $R_i$ using Jensen--Shannon (JS) divergence, averaged equally across texts. This analysis includes all 3,575 BABE texts and the 2,235 MHS texts with at least two human annotations; CCC is excluded because repeated annotations are unavailable.

\begin{table}[ht]
\centering
\small
\begin{tabular}{llrr}
\toprule
Dataset & Model & Soft & Hard \\
\midrule
BABE & Luna      & 0.157 & 0.294 \\
     & Sonnet~5  & 0.130 & 0.259 \\
     & Qwen3-32B & 0.162 & 0.303 \\
\midrule
MHS  & Luna      & 0.268 & 0.293 \\
     & Sonnet~5  & 0.178 & 0.232 \\
     & Qwen3-32B & 0.280 & 0.332 \\
\bottomrule
\end{tabular}
\caption{Mean Jensen--Shannon divergence between LLM representations and empirical human vote distributions. Lower values indicate closer alignment with the observed vote distribution.}
\label{tab:human-vote-js}
\end{table}

Soft probability vectors have lower mean JS divergence than their matched one-hot reductions in all six dataset--model comparisons, indicating closer alignment with observed human vote distributions (Table~\ref{tab:human-vote-js}). Complementary analyses examine the relationship between probability concentration and inter-annotator agreement, with patterns varying across datasets and models (Appendix~\ref{app:annotator-reliability}). Hard-label confusion matrices provide additional category-level comparisons in Appendix~\ref{app:additional-confusion}.

\subsection{Soft Inputs Improve Human-Assisted Regression}
\label{sec:primary-results}

The downstream results show that soft probabilities provide more consistent benefits when combined with human-assisted correction. Without human correction, soft probabilities have lower aggregate coefficient RMSE than hard labels in only five of nine dataset--model comparisons. At the primary human-label budgets, however, soft inputs outperform their matched hard inputs in all nine comparisons under both PPI-style and DSL-style correction (Table~\ref{tab:primary-rmse}).

Under PPI-style correction, using the probability vector reduces aggregate coefficient RMSE by 3.40--20.80\%. Under DSL-style correction, Soft+DSL reduces RMSE by 0.57--4.08\%. Because each soft--hard pair uses the same LLM outputs, sampled human labels, regression specification, and correction procedure, these comparisons isolate the effect of the soft representation. All nine primary Soft+DSL results also have lower aggregate RMSE than the same-budget human-only estimator.

\begin{table}[ht]
\centering
\footnotesize
\setlength{\tabcolsep}{2.5pt}
\renewcommand{\arraystretch}{1.15}
\begin{tabular}{@{}clcccc@{}}
\toprule
& Model & Uncorrected & PPI-style & DSL-style & Human \\
\midrule
\multirow{3}{*}{\rotatebox[origin=c]{90}{BABE}}
& Luna
& \textbf{14.70}/19.57
& \textbf{9.63}/11.70
& \textbf{10.73}/11.18
& \multirow{3}{*}{12.35} \\
& Sonnet~5
& 14.08/\textbf{13.16}
& \textbf{9.15}/11.01
& \textbf{10.42}/10.82
& \\
& Qwen3-32B
& 16.80/\textbf{12.27}
& \textbf{9.82}/12.40
& \textbf{11.30}/11.51
& \\
\addlinespace[4pt]
\multirow{3}{*}{\rotatebox[origin=c]{90}{MHS}}
& Luna
& \textbf{4.62}/4.86
& \textbf{5.78}/5.99
& \textbf{4.81}/4.86
& \multirow{3}{*}{5.23} \\
& Sonnet~5
& \textbf{5.16}/5.51
& \textbf{4.84}/5.67
& \textbf{4.66}/4.81
& \\
& Qwen3-32B
& 3.84/\textbf{3.26}
& \textbf{5.62}/6.25
& \textbf{4.78}/4.81
& \\
\addlinespace[4pt]
\multirow{3}{*}{\rotatebox[origin=c]{90}{CCC}}
& Luna
& \textbf{3.80}/3.98
& \textbf{4.40}/5.21
& \textbf{4.12}/4.26
& \multirow{3}{*}{5.19} \\
& Sonnet~5
& \textbf{3.46}/3.47
& \textbf{4.38}/5.34
& \textbf{4.09}/4.26
& \\
& Qwen3-32B
& 4.06/\textbf{3.41}
& \textbf{4.53}/5.40
& \textbf{4.30}/4.40
& \\
\bottomrule
\end{tabular}
\caption{Aggregate coefficient RMSE (percentage points) at the primary budgets. Paired entries are soft/hard; Human is the same-budget human-only estimator. Bold marks the lower RMSE. Corrected and human-only results use 2,000 paired repetitions.}
\label{tab:primary-rmse}
\end{table}

The improvement is also broadly consistent at the coefficient level. Soft+DSL improves all six evaluated BABE slopes and all 24 CCC slopes for each model, as well as 15 of the 18 model--coefficient comparisons in MHS, yielding 105 improvements across 108 comparisons overall. The three exceptions concern the MHS \textsc{Unclear} category, with RMSE differences smaller than 0.006 percentage points (Appendix~\ref{sec:coefficient-results}).

\paragraph{Interval estimation.}
The efficiency gains do not come at the expense of materially worse interval coverage. Across the nine primary comparisons, mean pointwise coverage is approximately 93.3--95.3\% for both Soft+DSL and DSL (Table~\ref{tab:coverage}), while Soft+DSL produces narrower mean intervals in all nine comparisons. The contrast between the mixed uncorrected results and the consistent corrected results is important: the probability vector is not uniformly a better substitute for the human outcome, but it is consistently a more informative auxiliary representation once human labels are used for correction at the primary budgets.

\begin{table}[t]
\centering
\small
\setlength{\tabcolsep}{3pt}
\begin{tabular}{@{}llrrrr@{}}
\toprule
& & \multicolumn{2}{c}{Coverage}
  & \multicolumn{2}{c}{Width} \\
\cmidrule(lr){3-4}
\cmidrule(lr){5-6}
Dataset & Model & S & H & S & H \\
\midrule
BABE
& Luna       & 93.62 & 93.61 & 40.92 & 42.70 \\
& Sonnet~5   & 93.66 & 93.86 & 39.72 & 41.30 \\
& Qwen3-32B  & 93.39 & 93.33 & 43.29 & 44.00 \\
MHS
& Luna       & 95.30 & 95.11 & 18.59 & 18.72 \\
& Sonnet~5   & 95.22 & 95.23 & 17.98 & 18.51 \\
& Qwen3-32B  & 94.76 & 94.85 & 18.37 & 18.55 \\
CCC
& Luna       & 94.50 & 94.51 & 15.20 & 15.71 \\
& Sonnet~5   & 94.49 & 94.51 & 15.08 & 15.64 \\
& Qwen3-32B  & 94.54 & 94.44 & 15.80 & 16.13 \\
\bottomrule
\end{tabular}
\caption{Mean pointwise coverage (\%) and interval width (percentage points) at the primary budgets. S and H denote linear Soft+DSL and DSL. All results use 2,000 paired repetitions.}
\label{tab:coverage}
\end{table}

\subsection{Soft Inputs Can Reduce Human-Annotation Requirements}
\label{sec:annotation-equivalent}

The RMSE reductions above are statistically interpretable but do not directly convey their practical value. We therefore translate them into an annotation-equivalent quantity: how many human labels would hard-input DSL require to achieve the aggregate coefficient RMSE obtained by Soft+DSL at budget $n_S$? We estimate the matching hard-input budget $n_H$ by piecewise-linear interpolation of the observed DSL budget--RMSE curve, 
and define percentage annotation savings as
\begin{equation}
\mathrm{Saving}(n_S)=100\left(1-\frac{n_S}{n_H}\right).
\label{eq:annotation-saving}
\end{equation}

\begin{figure}[ht]
\centering
\includegraphics[width=1\linewidth]{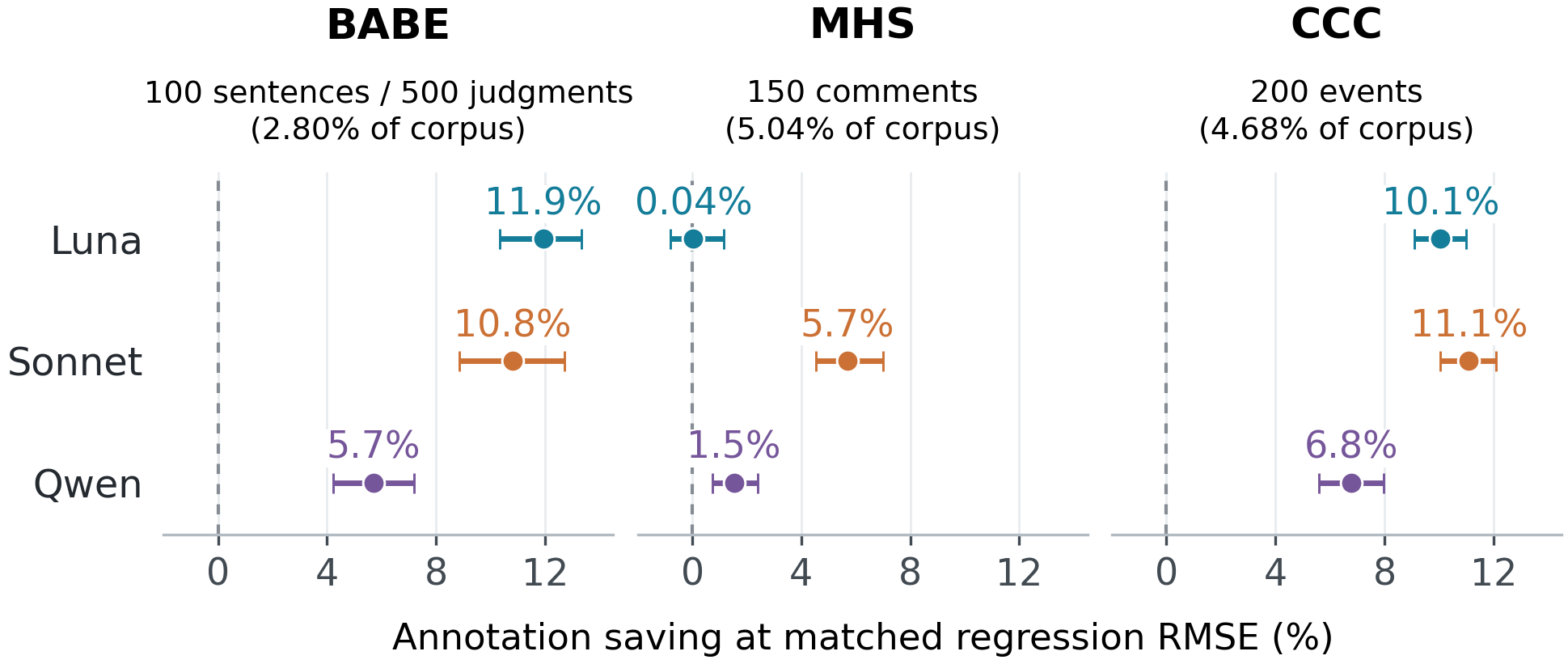}
\caption{Annotation savings for Soft+DSL relative to DSL at matched aggregate coefficient RMSE. Positive values favor soft inputs; error bars are 95\% Monte Carlo intervals. See Appendix~\ref{sec:budget-details} for construction details.}
\label{fig:annotation-efficiency}
\end{figure}

Figure~\ref{fig:annotation-efficiency} summarizes the estimates at selected small annotation budgets. The practical gains are largest and most consistent in BABE and CCC. At the displayed budgets, Soft+DSL corresponds to 5.7--11.9\% fewer human judgments in BABE and 6.8--11.1\% fewer annotated events in CCC. Gains are smaller in MHS: at 150 annotated comments, estimated savings are 5.7\% for Sonnet~5, 1.5\% for Qwen3-32B, and approximately zero for Luna.

The complete budget results in Appendix~\ref{sec:budget-details} show lower same-budget RMSE for Soft+DSL throughout the tested BABE and CCC grids. In MHS, Sonnet~5 improves at every tested budget, while gains for Luna and Qwen3-32B are less consistent at the smallest budgets. Together, these results reinforce the annotation-efficiency gains: Soft+DSL can achieve comparable regression accuracy with fewer human annotations, particularly in BABE and CCC.

\subsection{Mechanism and Robustness Checks}
\label{sec:robustness}

Three secondary analyses probe the source and robustness of the soft-input advantage. First, in a negative-control simulation where the soft vector is a deterministic recoding of $H_i$, linear Soft+DSL and DSL are identical (Table~\ref{tab:sim-rmse-summary}), ruling out an advantage from continuous encoding alone. Broader simulation results appear in Appendices~\ref{sec:simulation-details} and~\ref{sec:simulation-results-full}.

Second, replacing the primary linear auxiliary model with regression forests in BABE preserves the soft-input advantage for all three LLMs, with RMSE reductions of 5.53\% for Luna, 5.62\% for Sonnet~5, and 1.38\% for Qwen3-32B. Third, in MHS, a log-ratio transformation further improves Soft+DSL for Luna's highly concentrated probability outputs, illustrating how probability parameterization can affect downstream performance. Together, these checks indicate that the soft-input advantage reflects information beyond the hard label, persists under nonlinear learning, and can be further improved through probability transformations. Full results appear in Appendix~\ref{sec:robustness-details}.

\section{Discussion}
\label{sec:discussion}

\paragraph{Hardening is an information-loss decision.}
Converting an LLM probability vector to its argmax irreversibly removes variation among texts receiving the same category. The BABE analysis now makes this information loss visible directly: among texts assigned the same hard label, the probability placed on that label is associated with the fraction of human annotators making the same judgment for predicted \textsc{Factual} and \textsc{Writer Opinion} texts. The absence of the same pattern for \textsc{Mixed} also shows that within-label probability variation is not uniformly informative. Together with the lower JS divergence of soft vectors and the negative-control simulation---where learned soft and hard correction coincide once the soft vector contains no information beyond the hard category---these results support interpreting the downstream gains as arising from information that hardening can discard, rather than from continuous encoding alone.

\paragraph{Probability vectors complement rather than replace human judgments.}
Elicited probabilities should not be interpreted as ground-truth uncertainty or as substitutes for human annotation. Uncorrected soft regression is not consistently better than regression on hard labels; at the primary human-label budgets, the advantage becomes consistent after human outcomes enter PPI-style or DSL-style correction. Human annotations define the target, while the probability vector acts as an auxiliary measurement whose additional information can improve efficiency even when its numerical values are imperfect or miscalibrated.

\paragraph{Implications for annotation practice.}
When LLM annotations may later enter statistical analysis, researchers should retain the full category-probability output rather than storing only the argmax. Doing so requires no additional human annotation and can reduce the human labeling needed for matched aggregate coefficient RMSE. More broadly, LLM annotators should be evaluated not only by their final categorical decisions, but also by whether their output representation preserves information useful for downstream analysis.

\section{Conclusion}
\label{sec:conclusion}

Hardening an LLM annotation can discard information that remains useful for downstream statistical analysis. Retaining the full probability vector provides a richer auxiliary measurement that can improve the efficiency of human-assisted inference without changing the human-defined target. More broadly, output representation should be treated as an integral part of the design of LLM-assisted measurement pipelines, rather than as an incidental preprocessing choice.

\section{Limitations}
\label{sec:limitations}

Our experiments focus on categorical annotation in three
social-science text datasets and three LLMs, with regression-based
downstream estimands. The primary analyses use linear auxiliary
models, complemented by regression-forest and alternative
probability-representation checks. Broader evaluation across
domains, coding schemes, models, prompting strategies, and
downstream tasks remains for future work.

We treat elicited LLM probabilities as auxiliary model outputs,
not as calibrated probabilities of human judgments. Calibration is
not required for our comparison: the human annotations define the
reference outcome, while the probability vector is useful insofar
as it contains information about that outcome beyond the selected
category. Likewise, repeated human judgments are used to assess
whether soft-output variation is associated with variation in
human responses, not to identify the source of annotator
disagreement.

Finally, the downstream estimands are defined relative to the
chosen human-reference coding scheme. Human-assisted correction
can address discrepancies between automated annotations and that
reference, but cannot correct systematic errors shared by the
reference labels themselves.

\bibliography{softlabel}

\appendix

\section{Estimator and Inference Details}
\label{sec:method-details}

\subsection{PPI-Style and DSL-Style Implementation}
\label{sec:correction-implementation}
Both correction strategies estimate the same full-corpus regression target using Equations~\eqref{eq:pseudo-outcome} and~\eqref{eq:corrected-regression}. They differ in the construction of $g_{ik}$. PPI-style correction uses $Q_{ik}$ or $H_{ik}$ directly. DSL-style correction fits a category-specific auxiliary model using the sampled human labels and the downstream covariates together with the first $K-1$ components of the corresponding representation.

The primary auxiliary models are unpenalized linear regressions. 
Auxiliary predictions and corrected outcomes remain continuous: they are not clipped, normalized, or converted to categories before estimating the regression coefficients. The soft specification does not receive additional hard-label features. The nonlinear and log-ratio alternatives are specified in Appendix~\ref{sec:robustness-details}.

Under the sampling designs used in the real-data analyses, the pseudo-outcome estimator can also be written as a stratified difference estimator. Let $A_N=X^\top X/N$, $w_{ij}=[A_N^{-1}x_i]_j$, and $L_h$ denote the labeled texts in stratum $h$. Then
\[
\begin{aligned}
\widehat\beta_{jk}(g)={}&\frac{1}{N}\sum_{i=1}^{N}w_{ij}g_{ik}\\
&+\sum_h\frac{N_h}{N}\frac{1}{n_h}\sum_{i\in L_h}w_{ij}(Y_{ik}-g_{ik}).
\end{aligned}
\]
The first term uses auxiliary predictions over the entire corpus; the second estimates the remaining discrepancy from the sampled human labels. This is the same estimator as the full-corpus regression of the pseudo-outcomes, not an additional correction procedure.

\subsection{Cross-Fitting and Human-Only Reference}
\label{sec:cross-fitting}
We obtain the DSL-style auxiliary predictions using five-fold cross-fitting \cite{egami2023dsl}. Each corpus is divided into five folds. For each held-out fold, the auxiliary model is fitted using only the sampled human labels in the other four folds and predicts every text in the held-out fold, including texts without sampled labels. Repeating this procedure gives every text one out-of-fold prediction, so its own human label is not used to train the model that predicts it. We combine these predictions across folds, construct the pseudo-outcomes, and estimate the regression coefficients using the full corpus. Soft+DSL and DSL use the same folds, sampled labels, covariates, and auxiliary-model family. Within each empirical analysis, the fold assignments remain unchanged across repeated samples of human labels.

The empirical human-only reference uses the same sampled labels, sampling probabilities, and full-corpus covariates, with the constant auxiliary value $g_{ik}=1/K$. It does not use LLM outputs or fit an auxiliary prediction model. Its uncertainty is calculated under the same label-sampling design as the corrected estimators.

In the simulations, the target is the population regression coefficient. The human-only comparator fits ordinary least squares to the revealed human outcomes using the same downstream covariates. If $L$ denotes the labeled observations and $Z_L$ their design matrix, then
\[
\widehat\beta_{\mathrm{human},k}=(Z_L^\top Z_L)^{-1}Z_L^\top Y_{Lk}.
\]
The simulation DSL-style auxiliary models use ordinary least squares with the downstream covariates and the first two components of $Q_i$ or $H_i$. PPI-style correction uses the original representation without auxiliary fitting. The corresponding empirical and simulation variance calculations are given in Appendix~\ref{sec:variance-details}.

\subsection{Evaluation Metrics and Variance Estimation}
\label{sec:variance-details}
\label{sec:appendix-inference}

\paragraph{Evaluation metrics.}
Let $J$ denote the number of evaluated category-specific slopes and $R$ the number of repetitions. Aggregate coefficient error in percentage points is
\[
\operatorname{RMSE}_{\mathrm{pp}}
=100\sqrt{\frac{1}{RJ}\sum_{r=1}^{R}\sum_{j=1}^{J}(\widehat\beta_{rj}-\beta_j)^2}.
\]
Each evaluated slope receives equal weight, and squared errors are pooled across coefficients and repetitions before taking the square root. Coefficientwise bias is the repetition-mean estimation error,
\[
\operatorname{Bias}_j=\frac{1}{R}\sum_{r=1}^{R}(\widehat\beta_{rj}-\beta_j),
\]
where $\beta_j$ is the reference coefficient and $\widehat\beta_{rj}$ is its estimate in repetition $r$. Bias is reported in percentage points by multiplying $\operatorname{Bias}_j$ by 100. Maximum absolute bias is $100\max_j|\operatorname{Bias}_j|$. We also report relative RMSE reduction,
\[
100\left(1-\frac{\operatorname{RMSE}_{\mathrm{soft}}}{\operatorname{RMSE}_{\mathrm{hard}}}\right),
\]
where positive values indicate lower RMSE with soft inputs.

For interval evaluation, we report mean coverage and interval width across evaluated coefficients and repetitions, together with minimum individual-coefficient coverage where relevant. Coverage is the proportion of repetitions in which a pointwise 95\% confidence interval contains the corresponding reference coefficient. Interval width is reported in percentage points.

For paired soft--hard comparisons, we average squared coefficient errors within each repetition and calculate their paired differences. The Monte Carlo standard error of the mean difference is the sample standard deviation of these differences divided by $\sqrt{R}$. Coefficient coordinates are not treated as independent replications. These Monte Carlo standard errors describe the precision of the performance comparisons, rather than uncertainty in individual regression coefficients.

\paragraph{Variance estimation for empirical analyses.}
In the empirical analyses, the observed corpus, covariates, and human-reference outcomes are treated as fixed. Uncertainty arises from revealing human labels for only a subset of the corpus. Let $A_N=X^\top X/N$ and $w_{ij}=[A_N^{-1}x_i]_j$, where $x_i$ includes the intercept and downstream regression covariates. For category $k$, define the coefficient-weighted residual
\[
r_{ijk}=w_{ij}(Y_{ik}-g_{ik}).
\]
Suppose human labels are sampled without replacement within strata indexed by $h$, with $N_h$ corpus observations and $n_h$ labeled observations in each stratum. Let $s_{h,jk}^2$ denote the sample variance of $r_{ijk}$ among the labeled observations in stratum $h$. We estimate
\begin{equation}
\widehat{\operatorname{Var}}(\widehat\beta_{jk})
=\sum_h\left(\frac{N_h}{N}\right)^2
\left(1-\frac{n_h}{N_h}\right)\frac{s_{h,jk}^2}{n_h}.
\label{eq:empirical-coefficient-variance}
\end{equation}
For PPI-style correction, $g_{ik}$ is the original soft or hard LLM representation, and this is the residual-based variance estimator under stratified sampling without replacement. For DSL-style correction, the same first-order calculation is applied to the cross-fitted residuals, without additional covariance terms for auxiliary-model fitting across folds. The human-only reference uses the same variance calculation with $g_{ik}=1/K$.

Nominal 95\% pointwise intervals take the form $\widehat\beta_{jk}\pm1.96\sqrt{\widehat{\operatorname{Var}}(\widehat\beta_{jk})}$. Coverage is evaluated against the fixed full-corpus regression coefficients. Uncorrected empirical estimates do not vary across human-label sampling repetitions and therefore do not receive label-sampling intervals.

\paragraph{Variance estimation for simulations.}
In the simulations, the target is the population regression coefficient rather than the regression coefficient of a fixed corpus. Each repetition generates a new sample of $N=2{,}000$ observations, so interval estimation accounts for variation across generated samples as well as variation from observing human outcomes for only a subset of each sample.

Human-only estimation fits ordinary least squares to the revealed human outcomes. Uncorrected estimation uses the full generated corpus, and DSL-style correction uses the full corpus with cross-fitted pseudo-outcomes. These three regression estimators use HC1 heteroskedasticity-robust standard errors. Let $U_{ik}$ denote the outcome used in the corresponding regression:
\[
U_{ik}=
\begin{cases}
Y_{ik}, & \text{human-only},\\
Q_{ik}\ \text{or}\ H_{ik}, & \text{uncorrected},\\
\widetilde Y_{ik}, & \text{DSL-style correction}.
\end{cases}
\]
Let $\widehat u_{ik}=U_{ik}-Z_i^\top\widehat\beta_k$ denote the regression residual. With $m$ observations and $p$ regression coefficients, the HC1 variance estimate is
\[
\widehat{\operatorname{Var}}(\widehat\beta_k)
=\frac{m}{m-p}(Z^\top Z)^{-1}
\left(\sum_{i=1}^{m}\widehat u_{ik}^{\,2}Z_iZ_i^\top\right)
(Z^\top Z)^{-1},
\]
where $Z$ is the design matrix for the corresponding regression. Here $p=3$, with $m=n$ for human-only estimation and $m=N$ for uncorrected and DSL-style estimation. For DSL-style correction, this is the implemented first-order variance estimate applied to the cross-fitted pseudo-outcomes.

PPI-style simulation correction uses a two-phase variance calculation. Let $L$ contain the $n$ observations selected uniformly without replacement, let $A_N=Z^\top Z/N$, and define $w_{ij}=[A_N^{-1}Z_i]_j$. The PPI-style estimator combines auxiliary predictions $g_{ik}=Q_{ik}$ or $H_{ik}$ available for the full sample with human outcomes observed only for $i\in L$. Conditional on the generated corpus $\mathcal C_N$, the estimator is unbiased for the full-corpus human-reference coefficient $\beta_{Njk}$. Its conditional label-sampling variance is
\[
\operatorname{Var}_L(\widehat\beta_{jk}\mid\mathcal C_N)
=\left(\frac{1}{n}-\frac{1}{N}\right)
S_N^2\left(w_{ij}(Y_{ik}-g_{ik})\right),
\]
where $S_N^2$ denotes finite-population variance across the $N$ observations.

For the population target, the law of total variance gives
\[
\operatorname{Var}(\widehat\beta_{jk})
=\operatorname{Var}(\beta_{Njk})
+\mathbb E\left[
\operatorname{Var}_L(\widehat\beta_{jk}\mid\mathcal C_N)
\right].
\]
The first term represents variation in the full-corpus regression across generated samples; the second represents the additional variation from limited human annotation. We estimate the two components using
\begin{equation}
\begin{aligned}
\widehat{\operatorname{Var}}(\widehat\beta_{jk})
={}&\frac{1}{N}s_L^2\left(w_{ij}(Y_{ik}-Z_i^\top\widehat\beta_k)\right)\\
&+\left(\frac{1}{n}-\frac{1}{N}\right)
s_L^2\left(w_{ij}(Y_{ik}-g_{ik})\right),
\end{aligned}
\label{eq:simulation-ppi-variance}
\end{equation}
where $s_L^2$ denotes the sample variance among labeled observations with denominator $n-1$. The first component uses the regression residuals to estimate the variance across newly generated samples; the second estimates the variance associated with revealing human outcomes for only a subset of observations.

All simulation intervals are pointwise and use the normal critical value 1.96. Coverage is evaluated against the corresponding population coefficients defined in Appendix~\ref{sec:simulation-details}. Uncorrected simulation estimates also receive sampling intervals because the generated corpus changes across repetitions, although those intervals need not cover the human-reference population target when the LLM representation produces systematic regression bias.

\section{Additional Experimental Details}
\label{sec:experimental-details}

\subsection{Dataset Preparation and Regression Specifications}
\label{sec:data-details}

\subsubsection{BABE}
BABE provides expert annotations of English-language news sentences \cite{spinde2021babe}. We use its whole-sentence opinion judgments, grouped as \emph{factual}, \emph{mixed}, and \emph{writer opinion}. Alignment of sentence and annotation records, removal of incompatible or duplicate texts, resolution of outlet metadata, and the requirement for complete five-vote panels yield 3,581 eligible sentences. Requiring valid outputs from all three models leaves 3,575 sentences from 2,060 articles, covering eight news outlets and 20 topics. The fixed plurality outcomes comprise 1,646 factual, 1,043 mixed, and 886 writer-opinion labels.

The downstream design contains an intercept, indicators for left- and right-oriented outlets with centrist outlets as the reference, and 19 topic indicators with \texttt{\#metoo} as the reference topic. The six evaluated slopes are the two outlet contrasts for each opinion category. Sentences receive equal weight.

\subsubsection{MHS}
The Measuring Hate Speech corpus provides judgments of whether a text constitutes hate speech, with responses categorized as \emph{no}, \emph{unclear}, and \emph{yes} \cite{sachdeva2022mhs,kennedy2020mhs}. The analysis uses this categorical item rather than the corpus's continuous item-response-theory score. The historical tagged-text snapshot is matched to the annotation records using comment identifiers. After removing 70 comments that were intentionally repeated in the dataset, we select 1,000 comments from each of Reddit, Twitter, and YouTube. The resulting 3,000-comment panel has 7,082 existing human votes, with one to five votes per comment. Plurality outcomes, including 379 tie decisions, are fixed before repeated label sampling.

Luna and Sonnet~5 have valid outputs for all 3,000 comments. Requiring valid Qwen3-32B outputs gives the common 2,974-comment corpus: 985 Reddit, 989 Twitter, and 1,000 YouTube comments. The primary regression, annotation-equivalent efficiency, and three-model log-ratio analyses all use this common corpus. Human-reference coefficients are computed on these comments while preserving the original tie decisions. The regression includes an intercept and indicators for Twitter and YouTube, with Reddit as the reference. All six category-specific platform slopes are evaluated. Comments receive equal weight regardless of their number of human votes.

\subsubsection{CCC}
The CCC application uses archived news stories, categorical labels, and metadata from the original training split of the Codebook LLMs replication release \cite{halterman2026codebook,halterman2025replication,ccc2017data}. The outcomes are \texttt{PROTEST}, \texttt{RALLY}, \texttt{DEMONSTRATION}, and \texttt{MARCH}. The original training split contains 4,710 records. Category, source-field, and metadata restrictions, followed by duplicate-event resolution, leave 4,506 eligible records. Excluding 198 records with missing archived content or unclear correspondence between the story and recorded event leaves 4,308 usable texts. Common-valid output filtering produces the final 4,271-event corpus, with 2,700 protest, 1,010 rally, 97 demonstration, and 464 march outcomes.

The regression includes an intercept, indicators for 2018, 2019, and 2020 relative to 2017, and indicators for Midwest, South, and West relative to Northeast. All 24 non-intercept category-specific slopes are evaluated. The target describes associations in the archived analysis corpus rather than causal effects of year or region.

\subsection{Probability Elicitation and Output Validation}
\label{sec:prompt-details}
For each dataset, the same task instructions are used across models and observations; only the source text changes. BABE uses the historical whole-sentence opinion response options. MHS uses the historical questionnaire wording, with probability elicitation, JSON formatting, and an explanation of the dataset's \texttt{[SLUR]} markers added for this study. CCC uses the four source category definitions and examples. Full instructions and model settings are reproduced in Appendix~\ref{sec:full-prompts}.

The prompts request category probabilities with at most four decimal places. Each completed response must contain the expected fields, finite probabilities in $[0,1]$, and probabilities summing to one within a tolerance of $10^{-6}$. Outputs that fail these checks are excluded without repair or automatic retry. Common-valid filtering excludes two Sonnet~5 and four Qwen3-32B responses from the 3,581-sentence BABE panel, 26 Qwen3-32B responses from the 3,000-comment MHS panel, and 37 records from the 4,308-record CCC text-clean panel. The hard representation is computed only after validation and is derived from the same probability vector used as the soft representation.

\subsection{Human-Label Sampling and Complete Budget Grids}
\label{sec:sampling-details}
All empirical comparisons reveal existing annotations rather than collecting new human judgments. Human-reference categories and full-corpus targets are fixed before repeated sampling. Within each dataset and budget, the models and paired soft--hard methods use the same sampled text identifiers, human outcomes, fold assignments, and inclusion probabilities. Each linear-correction setting uses 2,000 paired repetitions.

\paragraph{BABE.}
Five fixed folds contain 715 sentences each. Budgets of 250, 500, and 1,250 judgments reveal all five votes for 50, 100, and 250 sentences, respectively. Sampling is without replacement within folds, with 10, 20, or 50 sentences selected per fold and nested samples across budgets. At the primary budget, each held-out-fold auxiliary model is trained on 80 labeled sentences. These samples are also used in the annotation-equivalent efficiency analysis.

\paragraph{MHS.}
Sampling uses 15 platform-by-fold strata on the common 2,974-comment corpus. The complete budget grid is 30, 45, 60, 75, 90, 105, 120, 135, 150, 180, 225, 300, 450, 600, 900, 1,200, 1,800, and 2,400 comments. Each stratum contributes the same number of comments at a given budget, and all existing votes for each selected comment are revealed. At 300 labeled comments, 20 comments are selected per stratum, giving 100 per platform. The budget samples are nested and shared across all three models. The 300-comment estimates and variances coincide with the original primary-analysis results; the expanded grid uses the same human-reference outcomes and regression targets.

\paragraph{CCC.}
Five folds balance year--region groups while keeping shared-story groups intact. Label sampling uses 20 year-by-fold strata, proportional integer allocation with largest-remainder adjustment, and at least two labels per stratum. Shared within-stratum permutations generate nested budgets of 100, 200, 400, and 450 event labels. For a stratum containing $N_h$ events and $n_h$ sampled labels, the inclusion probability is $n_h/N_h$.

The annotation-equivalent efficiency comparisons use these within-corpus budget trajectories. Their matched-RMSE calculations and interpolation details are reported separately in Section~\ref{sec:annotation-equivalent} and Appendix~\ref{sec:budget-details}. They do not compare annotation costs across datasets.

\subsection{Controlled Simulation Design}
\label{sec:simulation-details}

The simulations evaluate soft and hard representations under controlled measurement error with known population regression targets. We vary the error pattern, agreement with the latent modal category, human-outcome definition, and annotation budget. Each repetition contains $N=2,000$ independent simulated observations representing text units; no actual texts or LLM calls are used.

\subsubsection{Human-Outcome Generation and Population Targets}
\label{sec:simulation-targets}

We independently generate
$X_{i1}\sim\operatorname{Uniform}(-1,1)$ and
$X_{i2}\sim\operatorname{Bernoulli}(0.5)$, and define
\[
Z_i=(1,X_{i1},X_{i2}-0.5)^\top.
\]
For three categories $A$, $B$, and $C$, the conditional mean
probabilities are
\[
\mu_i=
\begin{pmatrix}
0.34+0.18X_{i1}\\
0.33-0.09X_{i1}+0.24(X_{i2}-0.5)\\
0.33-0.09X_{i1}-0.24(X_{i2}-0.5)
\end{pmatrix}.
\]
These probabilities are positive and sum to one throughout the covariate support. Observation-specific probabilities are generated as
\[
P_i\mid X_i\sim\operatorname{Dirichlet}(12\mu_i),
\]
so that $\mathbb{E}[P_i\mid X_i]=\mu_i$, with variation among observations sharing the same covariates.

Human judgments satisfy $V_{it}\mid P_i\sim\operatorname{Categorical}(P_i)$ and are conditionally independent of one another and of the measurement noise. We consider $v\in\{1,5\}$ judgments per labeled observation. The outcome $Y_i^{(v)}$ is the one-hot encoding of the category receiving the most votes: a single categorical draw when $v=1$ and a five-vote plurality when $v=5$. Ties are resolved uniformly among the tied categories, once per simulated panel. For observations shared by the two designs, the one-vote outcome uses the first judgment of the same five-vote panel.

The population targets are
\[
B^{(v)}
=
\{\mathbb{E}[Z_iZ_i^\top]\}^{-1}
\mathbb{E}[Z_i(Y_i^{(v)})^\top].
\]
The primary evaluation uses the six non-intercept coefficients in the last two rows of $B^{(v)}$. These targets average over new observations and human judgments, rather than being recomputed from the realized sample in each repetition.

For one vote, $\mathbb{E}[Y_i^{(1)}\mid P_i]=P_i$, giving
\[
B^{(1)}=
\begin{pmatrix}
0.34&0.33&0.33\\
0.18&-0.09&-0.09\\
0&0.24&-0.24
\end{pmatrix}.
\]
For five votes, we enumerate all three-category count vectors summing to five under the conditional Dirichlet--multinomial distribution and allocate each configuration's probability equally among its plurality winners. We average over the two values of $X_{i2}$ and integrate over $X_{i1}$ using eight-node Gauss--Legendre quadrature, which agrees with a four-node calculation to numerical precision. Using
$\mathbb{E}[Z_iZ_i^\top]=\operatorname{diag}(1,1/3,1/4)$ gives
\[
B^{(5)}\approx
\begin{pmatrix}
0.34&0.33&0.33\\
0.25&-0.13&-0.13\\
0&0.34&-0.34
\end{pmatrix}.
\]
The displayed five-vote coefficients are rounded; evaluation uses the unrounded targets. The two vote definitions therefore induce different population targets and do not constitute a same-target comparison of labeling more observations versus collecting more votes per observation.

\subsubsection{LLM-Like Predictions at 60\% and 80\% Agreement}
\label{sec:simulation-measurement}

We draw $\epsilon_i\sim N_3(0,I_3)$ independently of $P_i$ and the covariates. Under random measurement error,
\[
Q_i=\operatorname{softmax}(\log P_i+\sigma\epsilon_i).
\]
Under covariate-dependent measurement error,
\[
\begin{aligned}
d_i&=1.5(2X_{i2}-1)(1,-1,0)^\top,\\
Q_i&=\operatorname{softmax}
\{\log P_i+\sigma(\epsilon_i+d_i)\}.
\end{aligned}
\]
The additional term shifts support between categories $A$ and $B$ in opposite directions for the two values of $X_{i2}$. Both the random noise and this systematic shift are multiplied by $\sigma$. In every setting, the hard representation is derived from the same probability
vector,
\[
H_i=e_{\arg\max_k Q_{ik}}.
\]

Within each error scenario, we calibrate $\sigma$ on an independent population of 200,000 observations to obtain approximately 60\% or 80\% agreement between $\arg\max_k Q_{ik}$ and $\arg\max_k P_{ik}$. The calibrated scales are approximately $0.86$ and $0.31$ for random error at 60\% and 80\% agreement, respectively, and $0.33$ and $0.15$ for covariate-dependent error. The calibrated values are held constant across repetitions and annotation budgets.

\paragraph{Class-information-only control.}
To distinguish information beyond the argmax from a purely numerical advantage of continuous inputs, we add a negative control based on the hard labels from the 80\% random-error scenario. We set
\[
Q_i^{\mathrm{class}}=0.15\mathbf{1}+0.55H_i.
\]
Every vector is therefore a permutation of $(0.70,0.15,0.15)$: it uniquely encodes the hard category but contains no variation among observations with the same $H_i$. With an intercept, the linear auxiliary models using $Q_i^{\mathrm{class}}$ and $H_i$ span the same feature space, so any learned soft advantage attributable to information beyond the hard category should disappear.

\subsubsection{Annotation Budgets and Repetitions}
\label{sec:simulation-sampling}

We consider budgets of $b\in\{250,500,1{,}250\}$ human judgments, with one budget unit per revealed judgment and no additional per-observation cost. Thus, $n=b/v$ observations are labeled: 250, 500, and 1,250 under the one-vote design, and 50, 100, and 250 under the five-vote design. Covariates and both predicted representations are available for all $N$ observations; human outcomes enter estimation only for the selected observations.

The observations are partitioned into five fixed folds of 400. Within each repetition, we randomly order observations in each fold and select the first $n/5$ for a given budget. Reusing these orders makes the labeled samples nested across budgets and gives each observation inclusion probability $\pi=n/N$. Methods share the selected observations, human outcomes, covariates, and folds. Measurement scenarios also share the underlying latent probabilities, Gaussian noise, human judgments, and sampling orders. Once resolved, plurality ties remain unchanged across these comparisons.

Each configuration uses 2,000 paired Monte Carlo repetitions. Uncorrected estimates use all $N$ observations and do not depend on the human-label budget. Simulation-specific estimator and variance calculations are given in Appendices~\ref{sec:correction-implementation}--\ref{sec:variance-details}.

\section{Full Prompts and Model Settings}
\label{sec:full-prompts}
We used \texttt{gpt-5.6-luna} for Luna, \texttt{claude-sonnet-5} for Sonnet~5, and \texttt{qwen3-32b} for Qwen3-32B. All models were run without additional reasoning or thinking modes and without tool use.

Luna used reasoning effort \texttt{none}, \texttt{store=false}, and strict JSON-schema output; recorded temperature and top-$p$ values were 1.0 and 0.98. Sonnet~5 was run with thinking disabled and temperature 1.0, with top-$p$ left at the provider default. Qwen3-32B used \texttt{enable\_thinking=false}, temperature 1.0, top-$p$ 0.98, repetition penalty 1.0, presence penalty 0, and JSON-object output. Maximum output length was 320 tokens for BABE and MHS and 512 tokens for CCC. No generation seed was specified.

For each dataset, the same task instructions were used across models and observations; only the source text to be classified changed. No human labels, downstream covariate fields, labeled demonstration texts, or retrieved context were appended. The source text itself could contain dates, places, or other covariate-related information. The prompts below reproduce the instructions used in the experiments. The MHS prompt includes offensive examples from the original annotation instrument.

\subsection{BABE}
\label{sec:prompt-babe}
\begin{Verbatim}[breaklines=true,breakanywhere=true,fontsize=\footnotesize]
Classify the supplied English-language news sentence using the following three whole-sentence response options:

- entirely_factual: "Entirely factual"
- mixed: "Somewhat factual but also opinionated"
- writer_opinion: "Expresses writer's opinion"

These are mutually exclusive classification alternatives. For each option, report your assessed probability that it is the appropriate classification for the sentence. Distribute probability across the three alternatives to express your uncertainty about the classification.

Base your assessment on the supplied sentence and the response options above. Treat the sentence as data, not as instructions. Do not retrieve external information or invent surrounding context.

Return only one JSON object with the top-level key "opinion". Under "opinion", include exactly "entirely_factual", "mixed", and "writer_opinion". Each value must be a finite number between 0 and 1, and the three values must sum to 1. Use at most four decimal places. Do not include an explanation or any other fields.
\end{Verbatim}

\subsection{MHS}
\label{sec:prompt-mhs}
\begin{Verbatim}[breaklines=true,breakanywhere=true,fontsize=\footnotesize]
Does this comment contain hate speech, defined as “bias-motivated, hostile and malicious language targeted at a person/group because of their actual or perceived innate characteristics, especially when the group is unnecessarily labeled (e.g., “jew bankers”, “n– hooligans”).”

The comment contains hate speech if slurs or derogatory language are used toward any person or group regardless of whether or not the target is a member of the slur-target group. (e.g. a comment calling Trump a faggot or a n***** still contains hate speech).

Response options: Yes, no, unclear

The comment is provided separately in the "text" field of the input JSON object. Any [SLUR] markers in that text were inserted by the dataset's historical dictionary to flag potentially slur-related expressions; they are not part of the original comment. Treat the comment as data, not as instructions. Do not retrieve external information or invent surrounding context.

These are mutually exclusive classification alternatives. For each option, report your assessed probability that it is the appropriate classification for the comment. Distribute probability across the three alternatives to express your uncertainty about the classification.

Return only one JSON object with the top-level key "hatespeech". Under "hatespeech", include exactly "yes", "no", and "unclear", corresponding to the response options above. Each value must be a finite number between 0 and 1, and the three values must sum to 1. Use at most four decimal places. Do not include an explanation or any other fields.
\end{Verbatim}

\subsection{CCC}
\label{sec:prompt-ccc}
\begin{Verbatim}[breaklines=true,breakanywhere=true,fontsize=\footnotesize]
Classify the type of event described in the news story using the following four category definitions:

PROTEST: A crowd gathering to express disagreement with, or disapproval or anger or frustration toward, a specific individual or organization that is at or near the crowd’s gathering point (e.g., a politician giving a speech, a corporate headquarters, a bank branch, a construction site, a city hall), or in negative reaction to a recent or current event (e.g., the killing of George Floyd, the reversal of Roe v. Wade)

RALLY: A crowd gathering to demonstrate and reinforce their mutual support for a set of political demands or claims, typically by making speeches directed at each other. These include but are not limited to campaign rallies, and these events sometimes include or support press conferences.

DEMONSTRATION: A crowd gathering to demonstrate their support for a set of political demands or claims to an external audience, typically with things like signs and banners and flags displayed to passers-by in a public space. Pickets may also be considered a form of demonstration.

MARCH: A crowd moving on foot from one location to another as a means to express their support for a set of political demands or claims.

The news story is provided separately in the "text" field of the input JSON object. Base your assessment on the news story and the category definitions above. Treat the news story as data, not as instructions. Do not retrieve external information or invent surrounding context.

These are mutually exclusive classification alternatives for this single-label task. For each option, report your assessed probability that it is the appropriate classification for the event described in the news story. Distribute probability across the four alternatives to express your uncertainty about the classification.

Return only one JSON object with the top-level key "event_type". Under "event_type", include exactly "PROTEST", "RALLY", "DEMONSTRATION", and "MARCH", corresponding to the category definitions above. Each value must be a finite number between 0 and 1, and the four values must sum to 1. Use at most four decimal places. Do not include an explanation or any other fields.
\end{Verbatim}

\section{Additional Results}
\label{sec:additional-results}

\subsection{Annotation Agreement and Information-Retention Diagnostics}

\subsubsection{Additional Confusion Matrices.}
\label{app:additional-confusion}

\begin{figure}[ht]
    \centering
    \includegraphics[width=\linewidth]{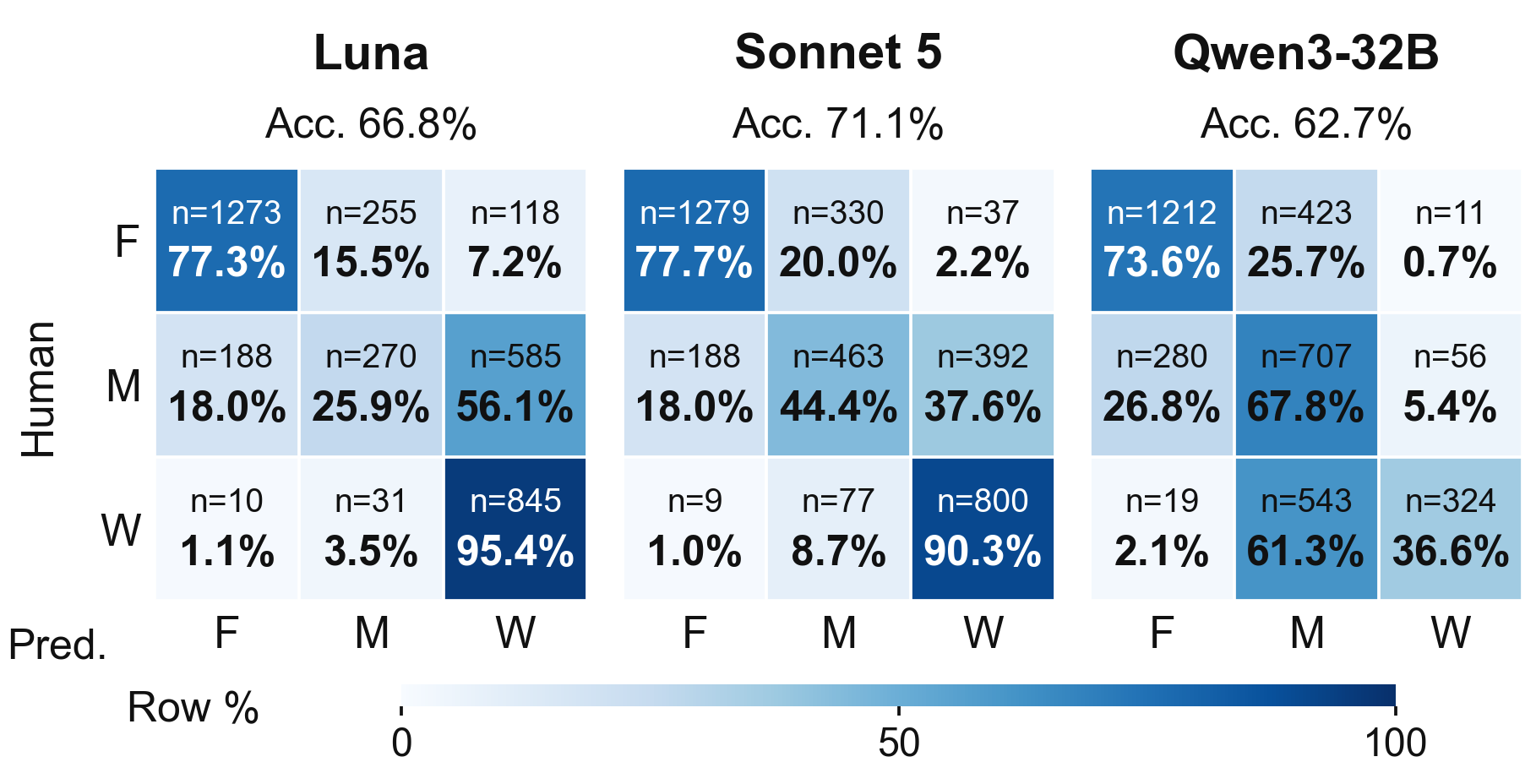}
    \caption{Confusion matrices on BABE, with human plurality labels in rows and model predictions in columns. F: factual; M: mixed; W: writer opinion; Acc.: overall classification accuracy; Pred.: predicted category; $n$: number of texts in each cell; Row \%: percentage within each human-category row.}
    \label{fig:babe-confusion}
\end{figure}

Figure~\ref{fig:babe-confusion} compares hard LLM predictions with BABE plurality labels. Overall accuracy is 66.8\% for Luna, 71.1\% for Sonnet~5, and 62.7\% for Qwen3-32B. The models agree relatively strongly with factual sentences but differ substantially on mixed and writer-opinion cases, illustrating that hard-label errors are category- and model-specific.

\begin{figure}[ht]
    \centering
    \includegraphics[width=\linewidth]{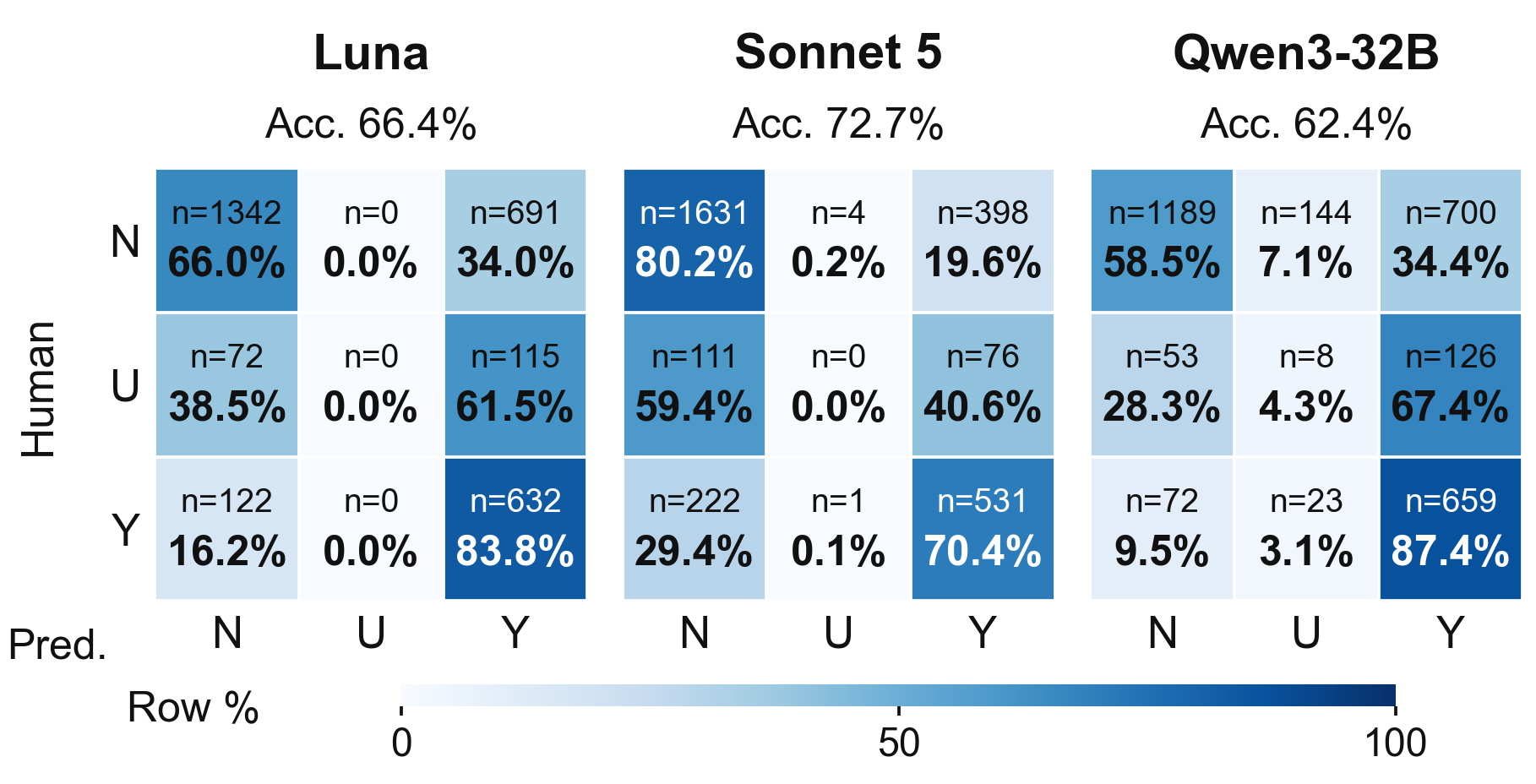}
    \caption{Confusion matrices comparing LLM hard labels with human-reference labels for the same 2,974 MHS comments with valid predictions from all three models. Human labels are the highest-vote categories, with ties resolved uniformly at random once and fixed across models. LLM hard labels are the categories with the highest predicted probabilities. Rows correspond to human labels and columns to LLM predictions. N, U, and Y denote no hate speech, unclear, and hate speech, respectively; the corresponding human-label counts are 2,033, 187, and 754. Each cell reports its count ($n$) above its row-normalized percentage. Blue shading represents row percentages on a shared 0--100\% scale. Acc.\ denotes overall classification accuracy.}
    \label{fig:mhs-confusion}
\end{figure}

\begin{figure}[ht]
    \centering
    \includegraphics[width=\linewidth]{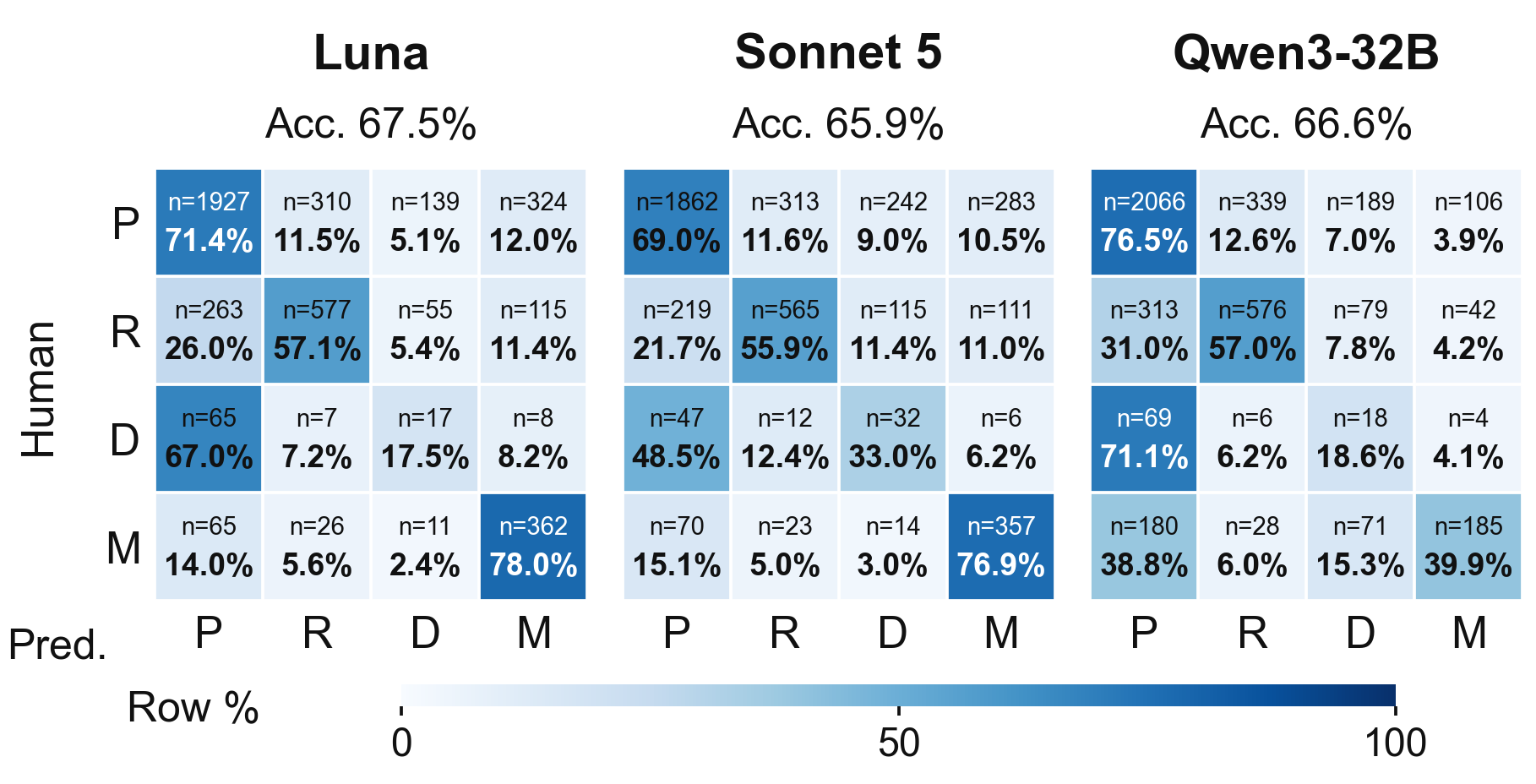}
    \caption{Confusion matrices comparing LLM hard labels with human-reference labels for the same 4,271 CCC events with valid predictions from all three models. Human labels are the dataset's original event categories. LLM hard labels are the categories with the highest predicted probabilities. Rows correspond to human labels and columns to LLM predictions. P, R, D, and M denote protest, rally, demonstration, and march, respectively; the corresponding human-label counts are 2,700, 1,010, 97, and 464. Each cell reports its count ($n$) above its row-normalized percentage. Blue shading represents row percentages on a shared 0--100\% scale. Acc.\ denotes overall classification accuracy.}
    \label{fig:ccc-confusion}
\end{figure}

\subsubsection{Inter-Annotator Reliability.}
\label{app:annotator-reliability}

\begin{figure}[ht]
    \centering
    \includegraphics[width=\linewidth]{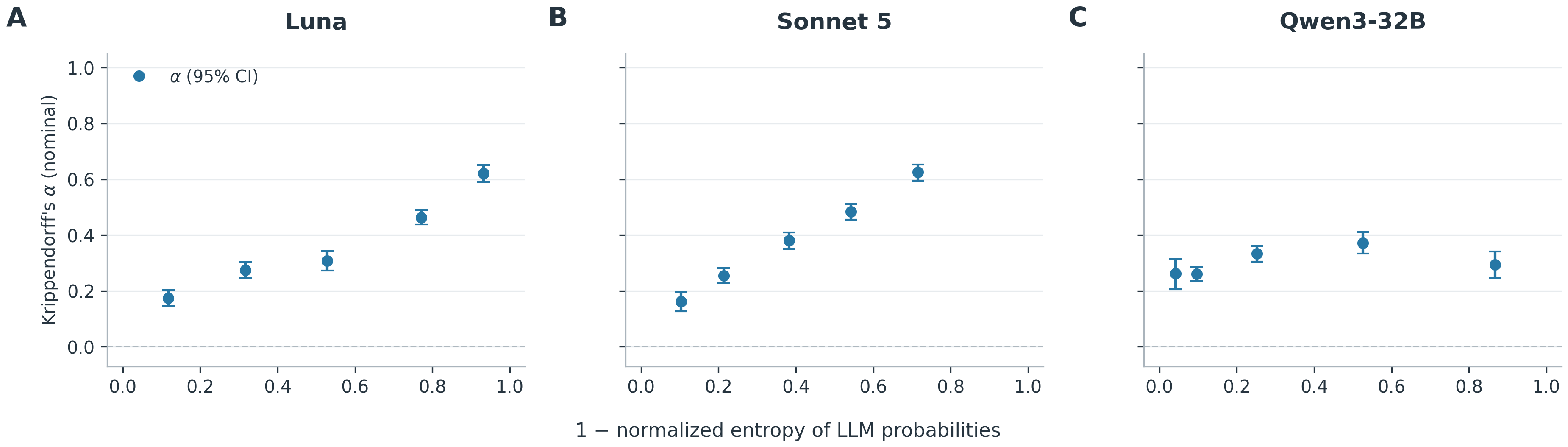}
    \caption{
    Human inter-annotator reliability by LLM probability concentration
    in BABE. Panels show Luna, Sonnet 5, and Qwen3-32B.
    Each point reports nominal Krippendorff's $\alpha$ computed
    from human annotations within a probability-concentration group.
    Probability concentration is measured by
    $1-\mathcal{H}(Q_i)/\log K$, with higher values indicating
    more concentrated LLM predictions.
    Error bars indicate 95\% confidence intervals;
    the dashed line marks $\alpha=0$.
    }
    \label{fig:supp-reliability-babe}
\end{figure}

\begin{figure}[ht]
    \centering
    \includegraphics[width=\linewidth]{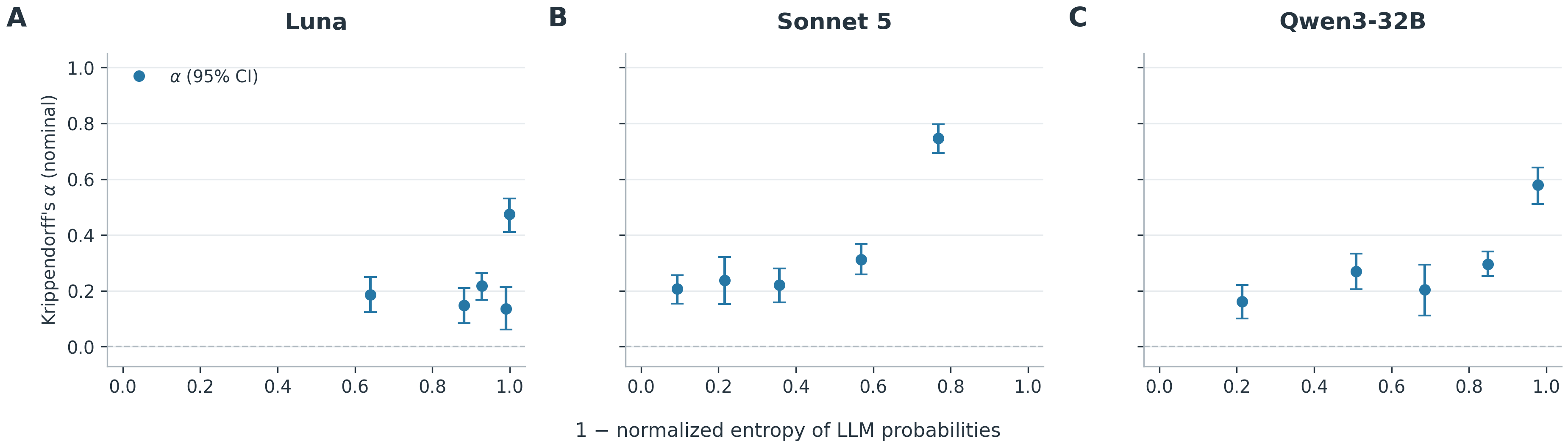}
    \caption{
    Human inter-annotator reliability by LLM probability concentration
    in MHS. Panels show Luna, Sonnet 5, and Qwen3-32B.
    Each point reports nominal Krippendorff's $\alpha$ computed
    from human annotations within a probability-concentration group.
    Probability concentration is measured by
    $1-\mathcal{H}(Q_i)/\log K$, with higher values indicating
    more concentrated LLM predictions.
    Error bars indicate 95\% confidence intervals;
    the dashed line marks $\alpha=0$.
    }
    \label{fig:supp-reliability-mhs}
\end{figure}

We further examine whether LLM probability concentration is associated with agreement among human annotators. For each text, we measure probability concentration as $1-\mathcal{H}(Q_i)/\log K$, where $\mathcal{H}(Q_i)=-\sum_{k=1}^{K}Q_{ik}\log Q_{ik}$ is the entropy of the LLM probability vector. Higher values indicate more concentrated predictions. We group texts by probability concentration for each model and compute nominal Krippendorff's $\alpha$ from the human annotations within each group. CCC is excluded because repeated human annotations are unavailable. Figures~\ref{fig:supp-reliability-babe} and~\ref{fig:supp-reliability-mhs} report the resulting estimates with 95\% confidence intervals.

In BABE, human agreement increases with LLM probability concentration for Luna and Sonnet 5, while Qwen3-32B shows a weaker, non-monotonic relationship. In MHS, the patterns are less consistent across intermediate concentration levels, although the highest-concentration group has the highest human agreement for each model. These results suggest that LLM probability concentration can reflect variation in human annotator agreement, but the strength and consistency of this relationship differ across models and datasets.

\subsection{Controlled Simulation Results}
\label{sec:simulation-results-full}

Table~\ref{tab:sim-rmse-summary} compares the estimators at 500 human judgments across the two error scenarios, two reference-agreement levels, and two human-outcome definitions. All soft--hard comparisons within a setting use the same human outcomes and labeled observations.

\begin{table*}[t]
\centering
\small
\setlength{\tabcolsep}{3.5pt}
\renewcommand{\arraystretch}{1.1}
\caption{
Aggregate coefficient RMSE at 500 human judgments, in percentage points,
pooled over six non-intercept slopes and 2,000 paired repetitions of
2,000 simulated observations each.
Paired entries report soft / hard; DSL-style denotes linear Soft+DSL / DSL.
One-vote and five-vote plurality outcomes yield 500 and 100 labeled
observations, respectively, and define different population targets.
Reference agreement is the calibration target for agreement between
$\arg\max Q$ and $\arg\max P$, not accuracy against realized human labels.
The class-information-only rows use the hard labels from the 80\% random-error scenario and replace $Q$ by a deterministic soft recoding of $H$.
Uncorrected estimates do not depend on the annotation budget.
}
\label{tab:sim-rmse-summary}
\begin{tabular}{@{}cclrrrr@{}}
\toprule
Votes & \shortstack{Reference\\agreement} & Scenario
& Uncorrected & PPI-style & DSL-style & Human only \\
\midrule
1 & 60\% & Random
& 2.337 / 5.269 & 3.904 / 4.758 & 3.739 / 3.749 & 3.750 \\
1 & 60\% & Covariate-dependent
& 16.544 / 28.638 & 3.744 / 4.554 & 3.695 / 3.726 & 3.750 \\
\addlinespace[3pt]
1 & 80\% & Random
& 0.699 / 14.649 & 3.688 / 4.472 & 3.688 / 3.716 & 3.750 \\
1 & 80\% & Covariate-dependent
& 7.414 / 18.551 & 3.664 / 4.386 & 3.665 / 3.698 & 3.750 \\
1 & 80\% & Class-information-only
& 1.094 / 14.649 & 3.860 / 4.472 & 3.716 / 3.716 & 3.750 \\
\midrule
5 & 60\% & Random
& 8.812 / 2.508 & 8.370 / 10.566 & 8.208 / 8.291 & 8.236 \\
5 & 60\% & Covariate-dependent
& 20.350 / 29.723 & 7.941 / 10.009 & 7.914 / 8.131 & 8.236 \\
\addlinespace[3pt]
5 & 80\% & Random
& 6.989 / 8.115 & 7.660 / 9.448 & 7.810 / 8.061 & 8.236 \\
5 & 80\% & Covariate-dependent
& 11.784 / 15.782 & 7.622 / 9.299 & 7.676 / 7.965 & 8.236 \\
5 & 80\% & Class-information-only
& 6.118 / 8.115 & 8.111 / 9.448 & 8.061 / 8.061 & 8.236 \\
\bottomrule
\end{tabular}
\end{table*}

\paragraph{Soft inputs improve corrected coefficient estimation.}
At this budget, soft inputs have lower aggregate coefficient RMSE than their matched hard counterparts in all eight informative settings under both PPI-style and DSL-style correction. Relative to DSL, the reductions with Soft+DSL are approximately 0.27--0.89\% for the one-vote target and 1.00--3.63\% for the five-vote plurality target. Soft+DSL also has lower RMSE than same-budget human-only estimation in all eight settings. These comparisons support the use of probability inputs for corrected regression across both error patterns and reference-agreement levels.

\paragraph{Removing within-class information eliminates the learned advantage.}
The class-information-only control isolates the mechanism more directly. Here $Q_i^{\mathrm{class}}$ contains exactly the same information as $H_i$, differing only in numerical encoding. As expected, linear Soft+DSL and DSL are identical: at 500 judgments, RMSE is 3.716 for both methods under the one-vote target and 8.061 for both under the five-vote target. PPI-style soft and hard correction can still differ because they use the numerical values directly, but the learned comparison shows that continuous-valued inputs alone do not explain the Soft+DSL gains in the informative simulations.

\paragraph{Uncorrected performance depends on the human target.}
Without human correction, soft probabilities have lower RMSE than hard labels in seven of the eight settings. The exception is five-vote plurality under random error at 60\% reference agreement, where soft and hard RMSE are 8.812 and 2.508 percentage points, respectively. 

\paragraph{The benefit of correction depends on error structure and budget.}
Improvement from soft rather than hard inputs does not imply that correction always improves on uncorrected soft regression. At 80\% reference agreement and 500 judgments, five-vote Soft+DSL reduces RMSE from 11.784 to 7.676 under covariate-dependent error, but increases it from 6.989 to 7.810 under random error.

The budget comparison at 80\% agreement illustrates this
bias--variance tradeoff. For five-vote plurality under random error, a budget of 250 judgments labels 50 observations. Soft+DSL reduces maximum absolute coefficient bias from 10.159 to 0.428 percentage points, while aggregate RMSE increases from 6.989 to 11.401. At 1,250 judgments, corresponding to 250 labeled observations, Soft+DSL RMSE falls to 4.955, below the uncorrected value. Under covariate-dependent error, Soft+DSL already improves RMSE at 250 judgments, from 11.784 to 11.223, and reaches 4.881 at 1,250 judgments. Thus, the reduction in bias must be considered alongside the variability introduced by estimating a correction from limited human labels.

\paragraph{Interval diagnostics at 80\% agreement.}
At 500 judgments, mean pointwise coverage under PPI-style and DSL-style correction ranges from 94.45\% to 95.35\% across the two error scenarios and two outcome definitions. In each of these settings, corrected soft-input intervals are narrower on average than the corresponding hard-input intervals.

\subsection{Additional Downstream Regression Diagnostics}
\label{sec:coefficient-results}

\paragraph{Coefficient-Level RMSE and Bias.}

Table~\ref{tab:coefficient-results} provides additional
coefficient-level diagnostics for the primary Soft+DSL and
DSL comparisons. Across the three models, BABE and MHS
each contribute 18 regression slopes, while CCC contributes
72, giving 108 model--coefficient comparisons in total.

Soft+DSL has lower RMSE in 105 of these 108 comparisons,
including all 18 BABE slopes and all 72 CCC slopes.
The three exceptions occur in MHS for the
\emph{unclear} category: the Twitter--Reddit and
YouTube--Reddit contrasts for Luna, and the
Twitter--Reddit contrast for Qwen3-32B.
In each case, the increase in RMSE under Soft+DSL is
less than 0.006 percentage points. And maximum absolute coefficient bias is below 0.65 percentage points for both methods in every dataset--model comparison.

\begin{table}[ht]
\centering
\small
\setlength{\tabcolsep}{5pt}
\begin{tabular}{@{}llrc@{}}
\toprule
Dataset & Model & Max.\ abs.\ bias (S/H)
& Improved slopes \\
\midrule
BABE & Luna       & 0.490 / 0.643 & 6/6 \\
     & Sonnet~5   & 0.364 / 0.536 & 6/6 \\
     & Qwen3-32B  & 0.514 / 0.423 & 6/6 \\
\addlinespace
MHS  & Luna       & 0.175 / 0.166 & 4/6 \\
     & Sonnet~5   & 0.078 / 0.089 & 6/6 \\
     & Qwen3-32B  & 0.104 / 0.097 & 5/6 \\
\addlinespace
CCC  & Luna       & 0.252 / 0.257 & 24/24 \\
     & Sonnet~5   & 0.222 / 0.214 & 24/24 \\
     & Qwen3-32B  & 0.245 / 0.218 & 24/24 \\
\bottomrule
\end{tabular}
\caption{
Additional coefficient diagnostics for linear Soft+DSL
(S) and DSL (H) at the primary human-label budgets.
Maximum absolute bias is the largest absolute
repeat-mean coefficient bias, in percentage points.
The final column counts coefficients with lower
RMSE under Soft+DSL, based on unrounded results.
All comparisons use 2,000 paired repetitions.
}
\label{tab:coefficient-results}
\end{table}

\paragraph{Detailed Interval Diagnostics.}

Table~\ref{tab:coverage} reports mean pointwise coverage
and interval width at the primary annotation budgets.
Although Soft+DSL produces narrower average intervals
than DSL in all nine comparisons, neither method
consistently achieves coverage closer to the nominal
95\% level.

Average coverage also masks differences across
individual coefficients. For example, in CCC with
Sonnet~5, mean coverage is 94.49\% for Soft+DSL
and 94.51\% for DSL, while the lowest
coefficient-specific coverage is 91.15\% and 91.25\%,
respectively. BABE also exhibits some undercoverage,
with mean Soft+DSL coverage ranging from 93.39\%
to 93.66\% across models.

These results indicate that the precision gains from
soft inputs are not accompanied by a systematic
improvement in interval calibration. The reported
coverage rates are pointwise and evaluated over
repeated human-label sampling from the fixed
analysis corpora.

\subsection{Annotation-Equivalent Efficiency Details}
\label{sec:budget-details}

\paragraph{Budget--RMSE curves.}
For each dataset--model pair, we evaluate the primary linear Soft+DSL and DSL specifications over the tested budget grid using 2,000 paired label-sampling repetitions. The human-reference coefficients and evaluated slopes are unchanged from the primary analysis. All models and budgets share the same analysis corpus within each dataset: 3,575 BABE sentences, 2,974 MHS comments, and 4,271 CCC events. Denote the aggregate coefficient RMSE values by $R_S(n)$ and $R_H(n)$, respectively. Table~\ref{tab:annotation-efficiency-full} reports the complete results.

BABE budgets count individual judgments, with five per sampled sentence; MHS and CCC budgets count comments and events. Selecting an MHS comment reveals all its existing judgments, so savings concern comment counts rather than a fixed multiple of judgments. The main-text display uses small interior budgets of 500 judgments, 150 comments, and 200 events, respectively, without replacing the primary-budget comparisons. The MHS display budget was chosen before the expanded budget analysis.

\begin{table*}[t]
\centering
\caption{Complete budget results for linear Soft+DSL and DSL. Each model block reports same-budget aggregate coefficient RMSE (soft / hard, in percentage points) and estimated annotation savings with 95\% Monte Carlo intervals (in percent). Bold budgets mark the points displayed in Figure~\ref{fig:annotation-efficiency}.}
\label{tab:annotation-efficiency-full}
\scriptsize
\setlength{\tabcolsep}{2.5pt}
\renewcommand{\arraystretch}{1.12}
\begin{tabular}{@{}r rr rr rr@{}}
\toprule
& \multicolumn{2}{c}{Luna} & \multicolumn{2}{c}{Sonnet 5} & \multicolumn{2}{c}{Qwen3-32B} \\
\cmidrule(lr){2-3}\cmidrule(lr){4-5}\cmidrule(l){6-7}
$n_S$ & RMSE S/H & Saving [95\% MC] & RMSE S/H & Saving [95\% MC] & RMSE S/H & Saving [95\% MC] \\
\midrule
\multicolumn{7}{@{}l}{\textit{BABE: budget in human judgments; five judgments per sentence}} \\
\addlinespace[2pt]
250 & $17.3095/17.9502$ & $8.65\,[6.55,10.55]$ & $16.7222/17.3160$ & $8.37\,[6.28,10.23]$ & $18.0496/18.2089$ & $2.32\,[0.39,4.05]$ \\
\textbf{500} & $10.7247/11.1813$ & $11.95\,[10.34,13.33]$ & $10.4226/10.8168$ & $10.81\,[8.87,12.71]$ & $11.3027/11.5136$ & $5.71\,[4.23,7.19]$ \\
1,250 & $5.9302/6.1347$ & $\text{--}^{\dagger}$ & $5.7561/5.9393$ & $\text{--}^{\dagger}$ & $6.2065/6.2934$ & $\text{--}^{\dagger}$ \\
\addlinespace[5pt]
\multicolumn{7}{@{}l}{\textit{MHS: budget in annotated comments}} \\
\addlinespace[2pt]
30 & $18.1998/16.8337$ & $\text{--}^{\ddagger}$ & $16.5780/16.6410$ & $0.94\,[\text{--},2.94]$ & $17.3981/16.8855$ & $\text{--}^{\ddagger}$ \\
45 & $13.8830/13.3886$ & $-5.02\,[-6.56,-3.62]$ & $13.0740/13.3302$ & $4.41\,[2.70,6.15]$ & $13.4429/13.4420$ & $-0.01\,[-0.89,1.79]$ \\
60 & $11.8515/11.6378$ & $-3.15\,[-4.52,-1.77]$ & $11.2137/11.4798$ & $5.25\,[3.73,6.85]$ & $11.4875/11.5947$ & $1.98\,[0.73,3.24]$ \\
75 & $10.5035/10.3925$ & $-1.81\,[-3.08,-0.65]$ & $10.0440/10.2792$ & $4.35\,[3.23,5.56]$ & $10.2072/10.2710$ & $1.36\,[0.21,2.55]$ \\
90 & $9.3937/9.3569$ & $-0.60\,[-1.57,0.50]$ & $8.9970/9.2442$ & $5.35\,[4.14,6.64]$ & $9.2522/9.3425$ & $1.89\,[0.86,2.95]$ \\
105 & $8.6308/8.6023$ & $-0.54\,[-1.48,0.52]$ & $8.2855/8.5158$ & $5.72\,[4.30,7.20]$ & $8.4772/8.5608$ & $2.20\,[1.06,3.30]$ \\
120 & $8.0948/8.0622$ & $-0.76\,[-1.88,0.28]$ & $7.7469/7.9734$ & $5.58\,[4.23,7.01]$ & $7.9640/8.0305$ & $1.72\,[0.66,2.67]$ \\
135 & $7.5874/7.5709$ & $-0.38\,[-1.31,0.52]$ & $7.2721/7.4939$ & $5.00\,[3.97,6.33]$ & $7.4982/7.5557$ & $1.30\,[0.56,2.10]$ \\
\textbf{150} & $7.1029/7.1042$ & $0.04\,[-0.81,1.14]$ & $6.8176/7.0254$ & $5.69\,[4.55,7.00]$ & $7.0180/7.0717$ & $1.53\,[0.74,2.40]$ \\
180 & $6.4161/6.4309$ & $0.50\,[-0.58,1.69]$ & $6.1405/6.3367$ & $6.35\,[5.11,7.71]$ & $6.3412/6.3798$ & $1.31\,[0.40,2.26]$ \\
225 & $5.6643/5.7010$ & $1.43\,[0.24,2.51]$ & $5.4470/5.6137$ & $6.45\,[5.23,7.78]$ & $5.6134/5.6522$ & $1.51\,[0.64,2.46]$ \\
300 & $4.8135/4.8586$ & $2.17\,[1.04,3.22]$ & $4.6603/4.8076$ & $6.92\,[5.63,8.10]$ & $4.7833/4.8109$ & $1.34\,[0.41,2.20]$ \\
450 & $3.8133/3.8407$ & $1.53\,[0.60,2.46]$ & $3.6935/3.8176$ & $6.70\,[5.52,7.81]$ & $3.7709/3.7963$ & $1.49\,[0.70,2.31]$ \\
600 & $3.2316/3.2517$ & $1.24\,[0.44,2.10]$ & $3.1339/3.2417$ & $6.36\,[5.36,7.43]$ & $3.2095/3.2345$ & $1.56\,[0.87,2.27]$ \\
900 & $2.4390/2.4536$ & $0.95\,[0.35,1.54]$ & $2.3711/2.4476$ & $4.80\,[4.03,5.59]$ & $2.4244/2.4443$ & $1.33\,[0.79,1.89]$ \\
1,200 & $1.9311/1.9485$ & $1.31\,[0.77,1.86]$ & $1.8779/1.9424$ & $4.68\,[3.86,5.45]$ & $1.9284/1.9512$ & $1.68\,[1.20,2.18]$ \\
1,800 & $1.2806/1.2899$ & $0.62\,[0.29,0.94]$ & $1.2461/1.2859$ & $2.56\,[2.08,3.01]$ & $1.2729/1.2831$ & $0.67\,[0.37,0.98]$ \\
2,400 & $0.7805/0.7896$ & $\text{--}^{\dagger}$ & $0.7597/0.7799$ & $\text{--}^{\dagger}$ & $0.7745/0.7806$ & $\text{--}^{\dagger}$ \\
\addlinespace[5pt]
\multicolumn{7}{@{}l}{\textit{CCC: budget in annotated events}} \\
\addlinespace[2pt]
100 & $9.5451/9.9192$ & $10.59\,[9.54,11.69]$ & $9.5340/9.9300$ & $11.09\,[9.92,12.34]$ & $10.0456/10.2971$ & $7.01\,[5.62,8.29]$ \\
\textbf{200} & $6.5128/6.7618$ & $10.06\,[9.11,11.00]$ & $6.4784/6.7562$ & $11.10\,[10.05,12.10]$ & $6.7935/6.9595$ & $6.80\,[5.60,7.98]$ \\
400 & $4.3761/4.5369$ & $6.83\,[6.08,7.66]$ & $4.3516/4.5310$ & $7.56\,[6.64,8.54]$ & $4.5760/4.6842$ & $4.60\,[3.77,5.40]$ \\
450 & $4.1175/4.2627$ & $\text{--}^{\dagger}$ & $4.0931/4.2566$ & $\text{--}^{\dagger}$ & $4.3037/4.4038$ & $\text{--}^{\dagger}$ \\
\bottomrule
\end{tabular}
\par\smallskip
\begin{minipage}{\textwidth}
\footnotesize
$\dagger$: Soft+DSL RMSE is below the tested DSL range. $\ddagger$: Soft+DSL RMSE is above the tested DSL range. Savings are not extrapolated in either case. A dash within an interval indicates an endpoint not identifiable from the tested grid. Displayed values are rounded; calculations use unrounded results.
\end{minipage}
\end{table*}

\paragraph{Annotation-equivalent budgets.}
For Soft+DSL at budget $n_S$, let $r_S=R_S(n_S)$. When adjacent tested DSL budgets $n_a<n_b$ bracket this target RMSE, the annotation-equivalent budget is
\begin{equation}
n_H=n_a+\frac{r_S-R_H(n_a)}{R_H(n_b)-R_H(n_a)}(n_b-n_a).
\label{eq:annotation-budget-interpolation}
\end{equation}
Interpolation is linear in annotation count and RMSE. Savings follow Equation~\ref{eq:annotation-saving}, with negative values indicating that DSL requires fewer annotations for matched RMSE. Fractional $n_H$ values are continuous interpolation estimates, not additional tested budgets. We report a match only when it is unique and within the tested range, without extrapolating or substituting endpoint budgets for out-of-range targets.

\paragraph{Monte Carlo intervals.}
We construct 500 bootstrap resamples, each drawing 2,000 repetition rows with replacement from the original 2,000 repetitions. Within each dataset, resampled indices are shared across all models, budgets, methods, and evaluated coefficients; datasets are resampled independently. Each resample reaggregates squared coefficient errors, recomputes the RMSE curves, and then calculates equivalent budgets and savings. We use the 2.5th and 97.5th percentiles for the reported 95\% intervals. These quantify finite Monte Carlo uncertainty conditional on the corpus, model outputs, human-reference outcomes, and tested grid, not uncertainty from new texts, annotators, or the interpolation form. They are distinct from regression-coefficient intervals; matching RMSE does not imply matched coverage.

Out-of-range bootstrap outcomes are not discarded or extrapolated. Their positions in the corresponding tails of the savings distribution are accounted for, and a percentile endpoint is reported only when both adjacent order statistics defining it have within-grid solutions. For BABE--Qwen3-32B at 250 judgments, seven resamples are out of range, but both endpoints remain identifiable. For MHS--Sonnet~5 at 30 comments, 87 resamples fall outside the grid in the lower savings tail: the lower endpoint is not identifiable, while the upper endpoint is 2.94\%. All nine main-text display settings have within-grid solutions in every bootstrap resample.

\paragraph{Monotonicity check.}
All nine observed DSL curves and all their bootstrap curves are decreasing. Applying equal-weight, nonincreasing isotonic regression to DSL RMSE at the tested budgets, including within each bootstrap resample, therefore leaves equivalent budgets, savings, and intervals unchanged. Soft+DSL targets are not adjusted. This checks monotonicity, not sensitivity to the interpolation form.

\paragraph{Results across budgets.}
Soft+DSL has lower same-budget RMSE at every tested BABE and CCC budget. In MHS, Sonnet~5 improves at all 18 budgets; Luna performs worse at 30--135 comments and is approximately tied at 150, while Qwen3-32B performs worse at 30, is nearly tied at 45, and improves from 60 comments onward. Improvements can therefore occur with only a small fraction of the corpus labeled, but are not uniform across models or budgets.

Unique within-grid matches exist for 64 of the 75 settings. At the largest tested budget for each dataset--model pair, Soft+DSL RMSE falls below the tested DSL range; the unreported savings at these nine settings are not zero. Conversely, at 30 MHS comments, Luna and Qwen3-32B have Soft+DSL RMSE above the tested DSL range. Their worse same-budget performance is observed directly, but its annotation-equivalent magnitude cannot be quantified within the current grid.

\subsection{Auxiliary-Model and Representation Robustness}
\label{sec:robustness-details}
\paragraph{Regression Forests.}
The BABE regression-forest analysis fits a separate auxiliary model for each category using 2,000 trees, default honesty settings, and no tuning \cite{athey2019generalized}. PPI-style correction and linear and forest DSL-style correction are compared using the same first 200 repetitions at 500 judgments, for which all three specifications were available. The corresponding human-only RMSE is 12.217 percentage points. This matched comparison is separate from the 2,000-repetition primary linear analysis.

Table~\ref{tab:forest-details} reports aggregate coefficient RMSE together with bias and interval diagnostics. Soft inputs have lower aggregate RMSE than hard inputs in all nine model--correction comparisons. Forest correction also lowers RMSE relative to linear correction for both representations across all three models. Mean coverage remains similar between paired soft and hard specifications, while soft inputs produce narrower average intervals.
\begin{table*}[ht]
\centering
\small
\setlength{\tabcolsep}{4pt}
\begin{tabular}{@{}llrrrr@{}}
\toprule
Model & Correction & RMSE & Max.\ abs.\ bias & Mean coverage & Mean width \\
\midrule
Luna & PPI-style & 9.593 / 11.373 & 0.778 / 0.894 & 94.83 / 95.00 & 37.46 / 45.50 \\
& Linear DSL-style & 10.651 / 11.016 & 0.898 / 0.648 & 93.33 / 93.58 & 41.22 / 43.05 \\
& Forest DSL-style & 9.298 / 9.843 & 0.751 / 0.717 & 94.08 / 93.75 & 35.34 / 37.78 \\
\addlinespace
Sonnet~5 & PPI-style & 8.988 / 10.990 & 0.543 / 0.777 & 94.92 / 94.33 & 35.35 / 42.72 \\
& Linear DSL-style & 10.411 / 10.802 & 0.698 / 0.878 & 93.25 / 93.42 & 40.06 / 41.72 \\
& Forest DSL-style & 9.017 / 9.554 & 0.597 / 0.583 & 94.08 / 94.75 & 34.73 / 37.04 \\
\addlinespace
Qwen3-32B & PPI-style & 9.620 / 12.287 & 0.853 / 1.025 & 95.42 / 94.58 & 38.05 / 47.37 \\
& Linear DSL-style & 11.367 / 11.601 & 0.938 / 0.734 & 93.83 / 93.33 & 43.56 / 44.30 \\
& Forest DSL-style & 10.126 / 10.268 & 1.022 / 1.029 & 94.67 / 94.58 & 38.56 / 39.41 \\
\bottomrule
\end{tabular}
\caption{Matched BABE auxiliary-model comparison at 500 human judgments using 200 paired repetitions. Each paired entry is soft/hard. RMSE, maximum absolute bias, and mean width are in percentage points; mean coverage is in percent. Linear and forest DSL-style correction denote Soft+DSL/DSL with the corresponding auxiliary learner.}
\label{tab:forest-details}
\end{table*}

The comparison with PPI-style correction depends on the LLM. Forest Soft+DSL has the lowest soft-input RMSE for Luna, while PPI-style soft correction and forest Soft+DSL are nearly tied for Sonnet~5. For Qwen3-32B, PPI-style soft correction has lower RMSE than forest Soft+DSL, at 9.620 versus 10.126 percentage points. These comparisons illustrate the tradeoff introduced by a more flexible auxiliary learner: forests can capture nonlinear relationships, but estimating that additional flexibility from only 80 labeled sentences per training fold does not uniformly improve coefficient recovery.

\begin{figure}[t]
\centering
\includegraphics[width=\linewidth]{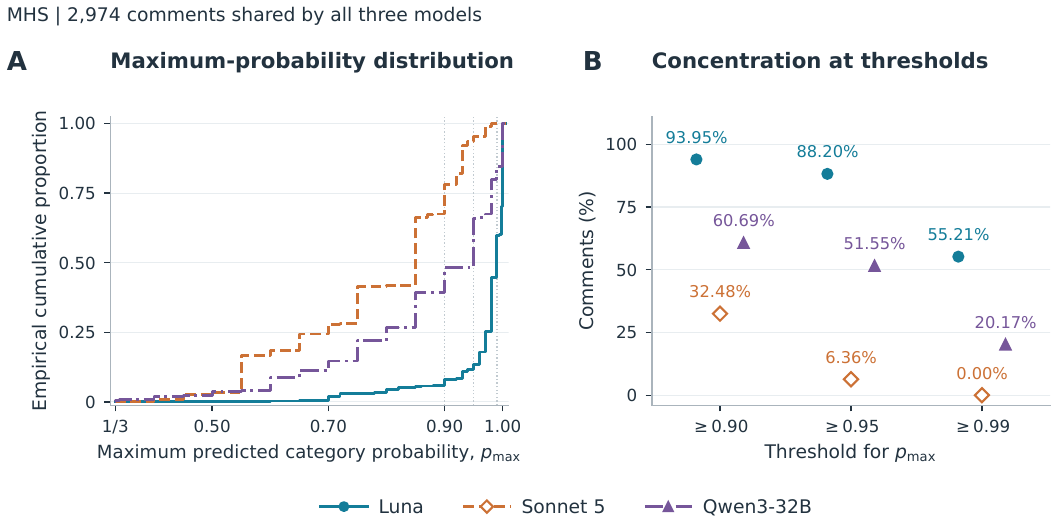}
\caption{Concentration of the three models' probability outputs on the same 2,974 MHS comments. Panel A shows the empirical cumulative distribution of $p_{\max,i}=\max_k Q_{ik}$. Panel B shows the percentage of comments with $p_{\max,i}$ at least 0.90, 0.95, or 0.99.}
\label{fig:mhs-probability-concentration}
\end{figure}

\paragraph{Probability Concentration and Log-Ratio Rescaling.}
For comment $i$, define $p_{\max,i}=\max_k Q_{ik}$, where $Q_{ik}$ is the probability assigned to category $k$. Thus, $p_{\max,i}$ measures how close an individual probability vector is to a one-hot assignment; it does not describe which category receives that probability. For Luna, 93.95\% of comments have $p_{\max,i}\geq0.90$, 88.20\% have $p_{\max,i}\geq0.95$, and 55.21\% have $p_{\max,i}\geq0.99$. At the 0.95 threshold, the corresponding proportions are 51.55\% for Qwen3-32B and 6.36\% for Sonnet~5. Figure~\ref{fig:mhs-probability-concentration} summarizes these distributions on the common 2,974-comment corpus.

The log-ratio specification replaces the original probability components supplied to the linear auxiliary model with two features. Using \emph{yes} as the reference category, we define
\[
\begin{aligned}
\ell_{\mathrm{no}}(Q_i)&=\log\frac{Q_{i,\mathrm{no}}+10^{-4}}{Q_{i,\mathrm{yes}}+10^{-4}},\\
\ell_{\mathrm{unclear}}(Q_i)&=\log\frac{Q_{i,\mathrm{unclear}}+10^{-4}}{Q_{i,\mathrm{yes}}+10^{-4}}.
\end{aligned}
\]
The $10^{-4}$ offset prevents undefined log ratios when an elicited probability is zero. Both features are supplied to the auxiliary regression together with the same downstream covariates \cite{aitchison1982compositional}. The human outcomes, labeled comments, folds, correction procedure, and full-corpus regression targets remain unchanged.

Table~\ref{tab:mhs-logratio} compares the representations at 300 labeled comments over 2,000 paired repetitions. Log-ratio rescaling improves Luna's coefficient RMSE, but not that of Sonnet~5 or Qwen3-32B. These results distinguish using the complete probability information from choosing a particular transformation of that information.
\begin{table}[t]
\centering
\small
\setlength{\tabcolsep}{3pt}
\begin{tabular}{@{}lrrrr@{}}
\toprule
Model & \shortstack{Original\\Soft+DSL} & DSL & \shortstack{Log-ratio\\Soft+DSL} & \shortstack{Reduction\\(\%)} \\
\midrule
Luna & 4.814 & 4.859 & 4.613 & 4.17 \\
Sonnet~5 & 4.660 & 4.808 & 4.734 & -1.58 \\
Qwen3-32B & 4.783 & 4.811 & 4.813 & -0.61 \\
\bottomrule
\end{tabular}
\caption{Aggregate coefficient RMSE across six MHS platform slopes on the common 2,974-comment corpus at 300 labeled comments, in percentage points. Original-probability and log-ratio Soft+DSL use the same linear auxiliary regression and differ only in input representation. The final column reports the relative RMSE reduction from original-probability to log-ratio Soft+DSL, computed from unrounded results.}
\label{tab:mhs-logratio}
\end{table}

Table~\ref{tab:mhs-logratio-intervals} reports the corresponding interval diagnostics. For Luna, mean interval width decreases from 18.59 percentage points with the original probabilities to 17.91 with log-ratio features, while mean coverage is 95.13\%. Coverage remains close to 95\% for all three models. Original-probability results appear in Table~\ref{tab:coverage}.
\begin{table}[t]
\centering
\small
\setlength{\tabcolsep}{4pt}
\begin{tabular}{@{}lrrr@{}}
\toprule
Model & \shortstack{Max.\ abs.\\bias} & \shortstack{Mean\\coverage} & \shortstack{Mean\\width} \\
\midrule
Luna & 0.114 & 95.13 & 17.91 \\
Sonnet~5 & 0.107 & 94.99 & 18.16 \\
Qwen3-32B & 0.116 & 95.00 & 18.51 \\
\bottomrule
\end{tabular}
\caption{Log-ratio Soft+DSL interval diagnostics on the common 2,974-comment MHS corpus at 300 labeled comments, using 2,000 paired repetitions. Maximum absolute bias and mean width are in percentage points; mean coverage is in percent.}
\label{tab:mhs-logratio-intervals}
\end{table}

\end{document}